\documentclass[preprint,12pt]{article}

\usepackage[T1]{fontenc}

\usepackage{latexsym}
\usepackage[dvipsnames]{xcolor}
\usepackage{oldlfont,rawfonts}
\usepackage{graphicx}
\usepackage{verbatim}

\usepackage{amssymb}
\usepackage{amsmath}
\usepackage{amsthm}

\usepackage{authblk}

\newcommand {\ent} {\mathrel{{\scriptstyle\mid\!\sim}}}

\newcommand {\sx} {\langle}
\newcommand {\dx} {\rangle}

\newcommand {\enne} {\mathcal{N}}

\newcommand {\tc} {\mid}
\newcommand {\vuoto} {\emptyset}

\newcommand{\tip}{{\bf T}}

\newcommand{\alc}{\mathcal{ALC}}
\newcommand{\lc}{\mathcal{LC}}

\newcommand{\alct}{\mathcal{ALC}+\tip}
\newcommand{\alctmin}{{\mathcal{ALC}}+\tip_{min}}
\newcommand{\el}{\mathcal{EL}^{\bot}}

\newcommand{\sroiqpt}{\mathcal{SROIQ}^{\Pe}\tip}

\newcommand{\alctr}{\mathcal{ALC}+\tip_{\textsf{\tiny R}}}

\newcommand{\alcFt}{\mathcal{ALC}^{\Fe}\tip}
\newcommand{\lcFt}{\mathcal{LC}^{\Fe}\tip}
\newcommand{\lcFtn}{{\mathcal{LC}}_{n}^{\Fe}\tip}

\newcommand{\lcnt}{{\mathcal{LC}}_{n}\tip}

\newcommand{\be}{\begin{enumerate}}
\newcommand{\ee}{\end{enumerate}}

\newcommand{\hide}[1]{}

\def \cases{\left \{\begin{array}{l}}
\def \endcases{\end{array}\right .}

\newcommand {\Fe} {{\bf F}}

\newcommand {\Pe} {{\bf P}}

\newcommand {\ri} {\rightarrow}
\newcommand {\Ri} {\Rightarrow}

\newcommand {\bes} {\begin{description}}
\newcommand{\ens} {\end{description}}
\newcommand {\la} {\langle}
\newcommand {\ra} {\rangle}

\newcommand {\beq} {\begin{quote}}
\newcommand {\enq} {\end{quote}}
\newcommand {\bit} {\begin{itemize}}
\newcommand {\enit} {\end{itemize}}

\newtheorem{lemma}{Lemma}
\newtheorem{corollary}{Corollary}
\newtheorem{proposition}{Proposition}
\newtheorem{definition}{Definition}
\newtheorem{example}{Example}

\begin{document}

\bibliographystyle{plain} 


\title{A preferential interpretation of MultiLayer Perceptrons in a conditional logic with typicality}

\author[1]{Mario Alviano} 
\author[2]{Francesco Bartoli} 
\author[2]{Marco Botta}

\author[2]{\\Roberto Esposito}  
\author[3]{Laura Giordano}  
\author[3]{\\Daniele Theseider Dupr{\'{e}}}  

\affil[1]{Universit\`{a} della Calabria, Italy}

\affil[2]{Universit\`{a} di Torino, Italy}

\affil[3]{Universit\`{a} del Piemonte Orientale, Italy}

\date{}

\maketitle

\begin{abstract}
In this paper we investigate the relationships between a multipreferential semantics for defeasible reasoning in knowledge representation and  a multilayer neural network model. Weighted knowledge bases for a simple description logic with typicality are considered under a (many-valued) ``concept-wise'' multipreference semantics. The semantics 
is used to provide a preferential interpretation of MultiLayer Perceptrons (MLPs).
A model checking and an entailment based approach are exploited in the verification of conditional properties of MLPs.  

\end{abstract}

%
%
%
%


\section{Introduction}

Preferential approaches to  commonsense reasoning 
\cite{Delgrande:87,Makinson88,Pearl:88,KrausLehmannMagidor:90,Pearl90,whatdoes,BenferhatIJCAI93,BoothParis98,Kern-Isberner01} 
have their roots in conditional logics \cite{Lewis:73,Nute80},
and  have been used to provide axiomatic foundations of non-monotonic or defeasible reasoning.
They have been extended to Description Logics (DLs) \cite{handbook}, to deal with inheritance with exceptions in ontologies,
by allowing for non-strict forms of inclusions,
called {\em typicality or defeasible inclusions}, 
with different preferential semantics \cite{lpar2007,sudafricaniKR,FI09},
and closure constructions \cite{casinistraccia2010,CasiniDL2013,AIJ15,Pensel18,CasiniStracciaM19,AIJ21,Casini21}.
Preferential extensions of DLs allow reasoning with exceptions through the identification of {\em prototypical properties} of individuals or classes of individuals.

In recent work, a ``concept-wise'' multi-preferential semantics has been proposed as a semantics of ranked knowledge bases (KBs) in a lightweight description logic \cite{TPLP2020}, in which defeasible or typicality inclusions of the form $\tip(C) \sqsubseteq D$ (meaning ``the typical $C$'s are $D$'s'' or ``normally $C$'s are $D$'s'') are given a rank, a natural number representing 
their strength. 
This two-valued concept-wise multi-preferential semantics, which takes into account preferences with respect to different concepts, 
has been shown to have some desirable properties from the knowledge representation point of view \cite{TPLP2020,NMR2020},
and has also been used to develop a preferential interpretation for Self-Organising Maps \cite{kohonen2001}, psychologically and biologically plausible neural network models.

The idea underlying the multi-preferential semantics is that different preferences should be associated to different concepts and, for instance, for two individuals Tom and Bob, and two concepts, {\em Swimmer} and {\em Student}, Tom might be more typical than Bob as a swimmer 
($\mathit{tom <_{Swimmer} bob}$) but less typical than $y$ as a student 
($\mathit{bob <_{Student} tom}$). 

In this paper, we focus on {\em weighted defeasible knowledge bases} (KBs), i.e., KBs in which typicality inclusions (conditionals) have a positive or negative {\em weight}, a real number representing the plausibility of the property. For instance, one may want to represent a situation in which students are normally young and use to have classes, while they usually do not have a scholarship.
In a weighted knowledge base these defeasible properties of students may be represented through some weighted typicality inclusions such as:

$\mathit{\tip(Student) \sqsubseteq Young, 80}$

$\mathit{\tip(Student) \sqsubseteq \exists hasClasses. \top, 90}$

$\mathit{\tip(Student) \sqsubseteq \exists hasScholarship. \top, -20}$

\noindent
where 
negative weights represent implausible properties, so that, in this example, it is rather implausible for students to have a scholarship, while it is quite plausible for them being young and having classes (with having classes slightly more plausible than being young).
Given such properties, a student Bob, who is young, has classes and has no scholarship, can be regarded as being more typical than a student Tom who is not young, but has classes and has a scholarship, so that $\mathit{bob  <_{Student} tom}$.
Similarly, for concept $\mathit{Swimmer}$ the prototypical elements can be characterized by a set of features and, hence, a set of typicality inclusions with their weights.
In our approach such features (such as being young or having classes) are as well represented as concepts in the description logic.

We do not assume that concepts are crisp, but that a domain element (Bob) may belong to a concept (e.g., Young) to some degree. Hence, we build our approach on fuzzy description logics, which have been widely studied in the literature (see, for instance, \cite{Straccia05,Stoilos05,LukasiewiczStraccia09,GarciaCerdanaAE2010,BorgwardtPenaloza12}). 
We develop a fuzzy formulation of the concept-wise multi-preferential semantics for weighted KBs in the description logic $\alc$, based on a {\em non-crisp} interpretation of typicality concepts (this different choice with respect to previous work, as we will see, has some impact on the properties of entailment).
We start from a fuzzy extension of $\alc$ \cite{LukasiewiczStraccia09},
and further extend it  with multiple preferences and with a non-crisp notion of typicality. The resulting fuzzy description logic with typicality is called $\alcFt$. To provide a semantics for weighted KBs, we introduce three different closure constructions for $\alcFt$, the {\em coherent}, the {\em faithful} and  the {\em $\varphi$-coherent} multi-preferential semantics for weighted knowledge bases.
Such constructions are similar in spirit to other semantic constructions adopted in the logics of commonsense reasoning, such as the {\em lexicographic closure} \cite{Lehmann95} and {\em c-representations} \cite{Kern-Isberner01,Kern-Isberner2014}, but  exploit multiple preference relations associated to concepts.

While similar (but different) semantic constructions for weighted knowledge bases have been considered in previous work based on different description 
logics\footnote{In particular, the coherent semantics was first introduced in \cite{JELIA2021} for weighted $\el$ KBs under a fuzzy semantics; the faithful semantics was considered in \cite{ECSQARU2021} for weighted $\alc$ KBs in the fuzzy case; the $\varphi$-coherent semantics was first proposed as an argumentation semantics in \cite{WorkshopAI3}. In all cases the interpretation of typicality was crisp.}, here we aim at a uniform formulation of the three semantics for $\alc$, under the assumption that the interpretation of typicality is non-crisp. This allows to study their mutual relationships, and to prove additional properties of multi-preferential entailment for the different semantics.

In particular, we show that any $\varphi$-coherent model of a weighted KB is a faithful (resp., coherent) model of the KB under suitable conditions, and that the notions of entailment under the different semantics satisfies (for some choice of fuzzy combination functions) all the {\em KLM properties  of a preferential consequence relation} \cite{KrausLehmannMagidor:90,whatdoes},
as well as other properties of the typicality operator studied in  \cite{FI09} for $\alct$, a two-valued typicality extension of  $\alc$. 
This contribution of the work extends the preliminary results on the properties
of the typicality logic 
investigated in \cite{ECSQARU2021} under the faithful semantics. 
In that case, the faithful semantics failed to satisfy all KLM properties of a preferential consequence relation, 
a negative result which is now overcome by adopting a non-crisp interpretation of typicality.
\normalcolor

The proposed (fuzzy) many-valued multi-preferential semantics are used in providing a {\em logical characterization} of Multilayer Perceptrons (MLPs) \cite{Haykin99}, which can be used for  post-hoc verification.
We will see that 
the input-output behavior of a multilayer network $\enne$ can be captured by a preferential interpretation $I_{\enne}^\Delta$ built over a set of input stimuli $\Delta$ (e.g., the training set), through a simple construction, which exploits the activity level of units for the input stimuli, thus allowing for the verification of properties of the network by {\em model checking} over the preferential interpretation.
We show that properties formalized as fuzzy typicality inclusions  in the boolean fragment of $\alcFt$ can be verified on the interpretation $I_{\enne}^\Delta$ in polynomial time in the size of $I_{\enne}^\Delta$ and in the size of the property. 
This is another contribution of the paper.

A logical characterization of a trained multi-layer networks $\enne$ is established by proving 
that the preferential interpretation $I_{\enne}^\Delta$, describing the network behavior over a set $\Delta$ of input stimuli, is indeed a $\varphi$-coherent model of the  weighted knowledge base $K^{\enne}$ and, vice-versa, that any $\varphi$-coherent model of the knowledge base $K^{\enne}$ captures the behavior of the network over some set $\Delta$ of input stimuli.
This strengthens the result in \cite{JELIA2021} that the interpretation $I_{\enne}^\Delta$ is a coherent model of $K^{\enne}$.

Undecidability results for fuzzy DLs with general inclusion axioms \cite{CeramiStraccia2013,BorgwardtPenaloza12} has led to consider a finitely-valued notion of the $\varphi$-coherent semantics, the $\varphi_n$-coherent semantics \cite{ICLP22}. 
In this paper, we prove that the $\varphi_n$ICLP22-coherent semantics is indeed an {\em approximation} of the $\varphi$-coherent semantics, which provides a full path from the definition of the fuzzy typicality logic with its semantics to the verification of properties of feedforward neural networks, which is based on proof methods developed for $\varphi_n$-coherent entailment \cite{arXiv2023} and on a Datalog encoding of the model checking approach 
\cite{Datalog2022}. 
In the experimentation, we exploit both the entailment-based approach and the model-checking approach in the verification of properties of trained multilayer feedforward networks. 
Such properties, expressed as fuzzy typicality inclusions, rely on typicality
in order to describe what the network has learned
to be a typical member of a class.
The experiments extend and complement the ones reported in \cite{ICLP22,Datalog2022,FCR2022}.

%

The schedule of the paper is the following.
Section   \ref{sec:ALC} contains the preliminaries about the description logic $\alc$ and its fuzzy version.
Section \ref{sec:fuzzyalc+T}
defines a (monotonic) extension of fuzzy $\alc$ (called $\alcFt$)
including a fuzzy notion of typicality. Section  \ref{sec:closure} introduces {\em weighted knowledge bases} 
in $\alcFt$ and their closure constructions through the notions of {\em faithful},  {\em coherent}  and  {\em  $\varphi$-coherent} (fuzzy) multi-preferential models. It establishes their relationships and defines the associated notions of entailment.
Section  \ref{sec:KLM} studies the properties of  $\alcFt$ and its closures, and proves that, for G\"odel  
fuzzy combination functions, $1$-entailment satisfies all KLM properties of a preferential consequence relation \cite{whatdoes} 
(properties that also extend to $k$-entailment, except for Cautious Monotonicity), while Rational Monotonicity does not hold. Further properties of the notion of typicality are also studied.
Section \ref{sec:multiulayer_perceptron} establishes the relationships between multi-preferential semantics and multilayer networks. It is proven that a multilayer network can be interpreted as a (fuzzy) multi-preferential interpretation (Section \ref{sec:fuzzy_sem_for_NN}), and that the network itself can be regarded as a weighted knowledge base in the boolean fragment of $\alcFt$ (Section  \ref{sec:NN&Conditionals}).
This allows both a {\em model-checking approach} and an {\em entailment approach} to be exploited for property verification.
Section \ref{sec:finitely_valued_case} proves that the $\varphi$-coherent models can be approximated in the finitely-valued case, which justifies the use of the $\varphi$-coherent semantics over a finite domain for verification. 
Section   \ref{sec:experimentation}  reports about experiments in the verification of properties of feedforward neural networks for the recognition of basic emotions, based on both the entailment and the model-checking approaches. 
The networks considered for entailment are significantly larger than the ones considered in \cite{TPLP2020,FCR2022}, implying a much larger search space for solving.
How model checking and entailment can be used together, and can be seen as complementary, is also pointed out.
Section \ref{sec:conclusions} concludes the paper with a discussion of related work and open issues.

%

The paper extends the work in \cite{Datalog2022,FCR2022} by substantiating it with several technical contributions, including
the above mentioned ones. 

\section{The description logic $\alc$ and fuzzy $\alc$}  \label{sec:ALC}

Fuzzy description logics have been widely studied in the literature for representing vagueness in description logics, e.g., by  \cite{Straccia05,Stoilos05,LukasiewiczStraccia09,BorgwardtPenaloza12},   
based on the idea that concepts and roles can be interpreted as fuzzy sets and fuzzy relations.
In fuzzy DLs, formulas have a truth degree from a truth space  $\cal S$, usually the interval $[0, 1]$, as  in Mathematical Fuzzy Logic \cite{Cintula2011}.
The finitely many-valued case is also well studied for DLs \cite{GarciaCerdanaAE2010,BobilloStraccia_InfSci_11,BobilloDelgadoStraccia2012,BorgwardtPenaloza13}.

In this section we recall the syntax and semantics of the description logic $\alc$ \cite{handbook} and of its fuzzy extension \cite{LukasiewiczStraccia09}.
We will also consider a finitely many-valued  
 fragment of $\alc$ with typicality.
 
 \subsection{$\alc$}

Let ${N_C}$ be a set of concept names, ${N_R}$ a set of role names
  and ${N_I}$ a set of individual names.  
The set  of $\alc$ \emph{concepts} (or, simply, concepts) can be
defined inductively  as follows:
 \begin{itemize}
\item
$A \in N_C$, $\top$ and $\bot$ are {concepts};
    
\item
if $C$ and $ D$ are concepts, and $r \in N_R$, 
then $C
\sqcap D,\; C \sqcup D,\; \neg C, \; \forall r.C,\; \exists r.C$ 
are {concepts}.
\end{itemize}

\noindent
A knowledge base (KB) $K$ is a pair $({\cal T}, {\cal A})$, where ${\cal T}$ is a TBox and
${\cal A}$ is an ABox.
The TBox ${\cal T}$ is  a set of concept inclusions (or subsumptions) $C \sqsubseteq D$, where $C,D$ are concepts.
The  ABox ${\cal A}$ is  a set of assertions of the form $C(a)$ and $r(a,b)$
where $C$ is a  concept, $a$ and $b$ are individual names in $N_I$ and $r$ a role name in $N_R$.

An  $\alc$ {\em interpretation}  is defined as a pair $I=\langle \Delta, \cdot^I \rangle$ where:
$\Delta$ is a domain---a set whose elements are denoted by $x, y, z, \dots$---and 
$\cdot^I$ is an extension function that maps each
concept name $C\in N_C$ to a set $C^I \subseteq  \Delta$, 
each role name $r \in N_R$ to  a binary relation $r^I \subseteq  \Delta \times  \Delta$,
and each individual name $a\in N_I$ to an element $a^I \in  \Delta$.
It is extended to complex concepts  as follows:

$\bot^I=\vuoto$, 

$\top^I=\Delta$,   

$(\neg C)^I=\Delta \backslash C^I$,

$(C \sqcap D)^I=C^I \cap D^I$,

$(C \sqcup D)^I=C^I \cup D^I$,

$(\exists r.C)^I =\{x \in \Delta \tc \exists y.(x,y) \in r^I \ \mbox{and}  \ y \in C^I\}$,   \ \ \ \ \ \ \  

$(\forall r.C)^I =\{x \in \Delta \tc \forall y. (x,y) \in r^I \Ri y \in C^I\}$.   \ \ \ \ \ \ \ \ \


\noindent
The notion of satisfiability of a KB  in an interpretation and the notion of entailment are defined as follows:

\begin{definition}[Satisfiability and entailment] \label{satisfiability}
Given an $\alc$ interpretation $I=\langle \Delta, \cdot^I \rangle$: 

	- $I$  satisfies an inclusion $C \sqsubseteq D$ if   $C^I \subseteq D^I$;
	
	-   $I$ satisfies an assertion $C(a)$ (resp., $r(a,b)$) if $a^I \in C^I$ (resp.,  $(a^I,b^I) \in r^I$).

\noindent
 Given  a knowledge base $K=({\cal T}, {\cal A})$, 
 an interpretation $I$  satisfies ${\cal T}$ (resp. ${\cal A}$) if $I$ satisfies all  inclusions in ${\cal T}$ (resp. all assertions in ${\cal A}$);
 $I$ is a \emph{model} of $K$ if $I$ satisfies ${\cal T}$ and ${\cal A}$.

 A subsumption $F= C \sqsubseteq D$ (resp., an assertion $C(a)$, $r(a,b)$),   {is entailed by $K$}, written $K \models F$, if for all models $I=$$\sx \Delta,  \cdot^I\dx$ of $K$,
$I$ satisfies $F$.

\end{definition}
Given a knowledge base $K$,
the {\em subsumption} problem is the problem of deciding whether an inclusion $C \sqsubseteq D$ is entailed by  $K$.
 
\subsection{Fuzzy $\alc$ and a finitely-valued $\alc$}

We shortly recall the semantics of a fuzzy extension of $\alc$, 
referring to the survey by Lukasiewicz and Straccia \cite{LukasiewiczStraccia09}.
We limit our consideration  
to a few features of a fuzzy DL  
and, in particular, we omit considering datatypes.

A {\em fuzzy interpretation} for $\alc$ is a pair $I=\langle \Delta, \cdot^I \rangle$ where:
$\Delta$ is a non-empty domain and 
$\cdot^I$ is {\em fuzzy interpretation function} that assigns to each
concept name $A\in N_C$ a function  $A^I :  \Delta \ri [0,1]$,
to each role name $r \in N_R$  a function  $r^I:   \Delta \times  \Delta \ri [0,1]$,
and to each individual name $a\in N_I$ an element $a^I \in  \Delta$.
A domain element $x \in \Delta$ 
belongs to the extension of $A$ to some degree in $[0, 1]$, i.e., $A^I$ is a fuzzy set.

The  interpretation function $\cdot^I$ is extended to complex concepts as follows: 

$\mbox{\ \ \ }$ $\top^I(x)=1$, $\mbox{\ \ \ \ \ \ \ \ \ \  \ \ \ \ \ }$ $\bot^I(x)=0$,   

 $\mbox{\ \ \ }$  $(\neg C)^I(x)= \ominus C^I(x)$,
 
 $\mbox{\ \ \ }$  $(C \sqcap D)^I(x) =C^I(x) \otimes D^I(x)$,  
 
 $\mbox{\ \ \ }$ $(C \sqcup D)^I(x) =C^I(x) \oplus D^I(x)$,

$\mbox{\ \ \ }$  $(\exists r.C)^I(x) = \sup_{y \in \Delta} \; r^I(x,y) \otimes C^I(y)$,  

$\mbox{\ \ \ }$  $(\forall r.C)^I (x) = \inf_{y \in \Delta} \;  r^I(x,y) \rhd C^I(y)$, 

\noindent
where  $x \in \Delta$, and $\otimes$, $\oplus$, $\rhd$ and $\ominus$ are 
arbitrary but fixed 
 triangular norm (or {\em t-norm}), triangular co-norm (or {\em s-norm}), 
implication function, and negation function, chosen among the combination functions of some fuzzy logic. 
In particular, in G\"odel logic $a \otimes b= min\{a,b\}$,  $a \oplus b= max\{a,b\}$,  $a \rhd b= 1$ {\em if} $a \leq b$ {\em and} $b$ {\em otherwise};
 $ \ominus a = 1$ {\em if} $a=0$  {\em and} $0$ {\em otherwise}. 
 In \L ukasiewicz logic, $a \otimes b= max\{a+b-1,0\}$,  $a \oplus b= min\{a+b,1\}$, 
 $a \rhd b= min\{1- a+b,1\}$ and $ \ominus a = 1-a$.
In Product Logic,  $a \otimes b= a \cdot b$,  $a \oplus b= a+b - a \cdot b$, 
 $a \rhd b= min\{1, b/a\}$ and  $ \ominus a = 1$ {\em if} $a=0$  {\em and} $0$ {\em otherwise}\footnote{Let us mention that any continuous t-norm can be expressed as an ordinal sum of copies of these three t-norms.}. 
Following \cite{LukasiewiczStraccia09}, we will not commit to a specific choice of combination functions,
 but in Tables \ref{propstsnorms} and \ref{propsimplneg} we report their main properties  (from Tables 1 and 2 in \cite{LukasiewiczStraccia09}).

\begin{table}
	\begin{footnotesize}
		\begin{tabular}{  c ||  c ||  c    }
			\hline
			\hline
			Axiom  &     T-norm   &  S-norm     \\	
			\hline
			\hline
			Tautology/contradiction 	&  $a \otimes 0	= 0$ 			&  $a \oplus 1 = 1 $ \\
			Identity 					&  $a \otimes 1	= a$ 			&  $a \oplus 0 = a $ \\
			Commutativity 				&  $a \otimes b	= b \otimes a $ &  $a \oplus b = b \oplus a $ \\
			Associativity 				&  $(a \otimes b) \otimes c = a \otimes (b \otimes c) $ &  $(a \oplus b) \oplus c = a \oplus (b \oplus c) $ \\
			Monotonicity				&  if $b \leq c$, then $ a \otimes b \leq a \otimes c$ &  if $b \leq c$, then $ a \oplus b \leq a \oplus c$ \\
			\hline
			\hline
		\end{tabular}
	\end{footnotesize}
	\caption{\label{propstsnorms} Properties for t-norms and s-norms}
\end{table}

\begin{table}
	\begin{footnotesize}
		\begin{tabular}{   c || c ||  c   }
			\hline
			\hline
			Axiom  &     Implication function   &  Negation function     \\	
			\hline
			\hline
			Tautology/contradiction 	&  $ 0 \rhd b = 1, a \rhd 1 = 1, 1 \rhd 0 = 0 $ 	&  $\ominus 0 = 1, \ominus 1 = 0 $ \\
			Antitonicity 				&   if $a \leq b$, then $ a \rhd c \geq b \rhd c$ 			& if $a \leq b$, then $ \ominus a \geq \ominus b $ \\
			Monotonicity				&  if $b \leq c$, then $ a \rhd b \leq a \rhd c$ &   \\
			\hline
			\hline
		\end{tabular}
	\end{footnotesize}
	\caption{\label{propsimplneg} Properties for implication and negation functions}
\end{table}

The  interpretation function $\cdot^I$ is also extended  to non-fuzzy axioms (i.e., to strict inclusions and assertions of an $\alc$ knowledge base) as follows:

$\mbox{\ \ \ }$   $(C \sqsubseteq D)^I= \inf_{x \in \Delta}  C^I(x) \rhd D^I(x)$
 
$\mbox{\ \ \ }$  $(C(a))^I=C^I(a^I)$

$\mbox{\ \ \ }$  $(R(a,b))^I=R^I(a^I,b^I)$.

A {\em fuzzy $\alc$ knowledge base} $K$ is a pair $({\cal T}, {\cal A})$ where ${\cal T}$ is a fuzzy TBox  and ${\cal A}$ a fuzzy ABox. A fuzzy TBox is a set of {\em fuzzy concept inclusions} of the form $C \sqsubseteq D \;\theta\; n$, where $C \sqsubseteq D$ is an $\alc$ concept inclusion axiom, $\theta \in \{\geq,\leq,>,<\}$ and $n \in [0,1]$. A fuzzy ABox ${\cal A}$ is a set of {\em fuzzy assertions} of the form $C(a) \theta n$ or $r(a,b) \theta n$, where $C$ is an $\alc$ concept, $r\in N_R$, $a,b \in N_I$,  $\theta \in \{{\geq,}\leq,>,<\}$ and $n \in [0,1]$.
Following Bobillo and Straccia  \cite{BobilloOWL2EL2018}, we assume that fuzzy interpretations are {\em witnessed}, 
i.e., the sup and inf are attained at some point of the involved domain.

We refer to fuzzy concept inclusions and fuzzy assertions as {\em fuzzy axioms}.

\begin{example} \label{Exa1}
Let us consider the fuzzy concepts $\mathit{Tall}$ (tall individuals) and $\mathit{\exists hasFriend.Tall}$  (the individuals having a tall friend), where hasFriend might as well be a fuzzy role, as a domain individual Bob may be friend of Mary to a given degree, e.g., $\mathit{ hasFriend^I(bob^I,mary^I)= 0.7}$.
Similarly, we may consider the fuzzy concept $\mathit{\exists hasParent.Tall}$.

For instance, we may have fuzzy assertions such as $\mathit{ hasFriend(bob,mary) }$  $\mathit{ \geq 0.5}$ or $\mathit{ Tall(mary)\geq 0.8}$ in the ABox ${\cal A}_f$, and 
fuzzy concept inclusions such as
$\mathit{\exists hasParent.Tall \sqsubseteq Tall \geq 0.7}$ (an individual having at least a tall parent is tall, holding to a degree greater than 0.7)
or
$\mathit{\forall hasFriend.Nerd \sqsubseteq Nerd}$ $\mathit{ \geq 0.8}$ (an individual having all nerd friends is a nerd, holding  to a degree greater than 0.8)
in the TBox ${\cal T}_f$.

Let us assume G\"odel logic, and that Bob 
has parents Mary and Tom. 
Consider an interpretation $I$ such that:
\begin{quote}
$\mathit{ hasParent^I(bob^I,mary^I)=1}$, 
$\mathit{ hasParent^I(bob^I,tom^I)= 1}$, \\
$\mathit{ hasParent^I(bob^I,z)= 0}$, for all $z \in \Delta$ with $z \neq mary^I, tom^I$,\\
$\mathit{ hasParent^I(x,z)= 0}$, for all  $x,z \in \Delta$ with $x \neq bob^I$,\\
$\mathit{ Tall^I(bob^I)= 0.8}$,
$\mathit{ Tall^I(mary^I)= 0.5}$,
$\mathit{ Tall^I(tom^I)= 0.9}$,\\
$\mathit{ Tall^I(x)= 0.5}$, for all $x \in \Delta$ with $x \neq bob^I, mary^I, tom^I$.
\end{quote}
As an example, we show that  $\mathit{(\exists hasParent.Tall \sqsubseteq Tall )^I = 0.8}$ holds.
We have to show that 
$\mathit{inf_{x \in \Delta} (\exists hasParent.Tall)^I(x) \rhd Tall^I(x) = 0.8}$.

\noindent
Note that:

$\mathit{ (\exists hasParent.Tall)^I(x) = \sup_{y \in \Delta} \; hasParent^I(x,y) \otimes Tall^I(y) }$ 

\ \ \ \ \ \ \ \ \ \ \ \ \ \ \ \ \ \ \ \ \ \ \ \ \ \ \ \ $\mathit{= \sup_{y \in \Delta} \; min\{hasParent^I(x,y), Tall^I(y)\} }$,  

\noindent
In particular, for $\mathit{ x= bob^I}$  we have:

$\mathit{(\exists hasParent.Tall)^I(bob^I) = \sup_{y \in \Delta} \; min\{hasParent^I(bob^I,y), Tall^I(y)\} }$,

\ \ \ \ \ \ \ \ \ \ \ \ \ \ \ \ \ \ \ \ \ \ \ \ \ \ \ \  $\mathit{=  \sup_{y \in \Delta} \; min\{hasParent^I(bob^I,y), Tall^I(y)\} = 0.9}$,

\noindent
In fact, we have three possible cases for $y$, $y=mary^I$, $y = tom^I$ and $y \neq mary^I, tom^I$:

$\mathit{ min\{hasParent^I(bob^I,mary^I), Tall^I(mary^I)\} = min\{1, 0.5\}= 0.5}$  

$\mathit{ min\{hasParent^I(bob^I,tom^I), Tall^I(tom^I)\} = min\{1, 0.9\}= 0.9}$ 

$\mathit{ min\{hasParent^I(bob^I,z), Tall^I(y)\} = min\{0, Tall^I(z)\}= 0}$, 

\noindent
for $y \neq mary^I, tom^I$. We take the maximum among the values,
then

$\mathit{(\exists hasParent.Tall)^I(bob^I) \rhd Tall^I(bob^I) = 0.9 \rhd 0.8=0.8}$.

\noindent
In a similar way, one can see that, for all $x\neq bob^I$,

$\mathit{(\exists hasParent.Tall)^I(x) \rhd Tall^I(x) = 0 \rhd 0.5=1}$,

\noindent
where $\mathit{(\exists hasParent.Tall)^I(x) =0}$ as, for $x \neq bob^I$,  $\mathit{ hasParent^I(x,z)= 0}$  for all  $z \in \Delta$.

Thus:
$\mathit{inf_{x \in \Delta} (\exists hasParent.Tall)^I(x) \rhd Tall^I(x) = 0.8}$.

\end{example}

The notions of satisfiability of a KB  in a fuzzy interpretation and of entailment are defined in the natural way.
\begin{definition}[Satisfiability and entailment for fuzzy KBs] \label{satisfiability-fuzzy}
A  fuzzy interpretation $I$ satisfies a fuzzy $\alc$ axiom $E$ (denoted $I \models E$), as follows:

- $I$ satisfies a fuzzy  inclusion axiom $C \sqsubseteq D \;\theta\; n$ if $(C \sqsubseteq D)^I \theta\; n$;

- $I$ satisfies a fuzzy  assertion $C(a) \; \theta \; n$ if $C^I(a^I) \theta\; n$;
 
- $I$ satisfies a fuzzy  assertion $r(a,b) \; \theta \; n$ if $r^I(a^I,b^I) \theta\; n$,

\noindent
where $\theta \in \{\geq,\leq,>,<\}$.

\noindent
Given  a fuzzy $\alc$ knowledge base $K=({\cal T}, {\cal A})$,
 a fuzzy interpretation $I$  satisfies ${\cal T}$ (resp. ${\cal A}$) if $I$ satisfies all fuzzy  inclusions in ${\cal T}$ (resp. all fuzzy assertions in ${\cal A}$).
A fuzzy interpretation $I$ is a \emph{model} of $K$ if $I$ satisfies ${\cal T}$ and ${\cal A}$.
A fuzzy axiom $E$   is {\em  entailed} by a fuzzy knowledge base $K$, written $K \models E$, if for all models $I=$$\sx \Delta,  \cdot^I\dx$ of $K$,
$I$ satisfies $E$.
\end{definition}
 
\begin{example}
Referring to the interpretation $I$ in Example \ref{Exa1},  we have seen that 
$\mathit{(\exists hasParent.Tall \sqsubseteq Tall )^I = 0.8}$ holds. Then, we can conclude that 
axiom $\mathit{\exists hasParent.Tall \sqsubseteq Tall \geq 0.7}$ is satisfied in $I$.
\end{example}

For the finitely many-valued case, we restrict to the boolean fragment $\lc$ of $\alc$ with no roles (and no universal and existential restrictions). 
We assume the truth space to be  
${\cal C}_n= \{0, \frac{1}{n},\ldots,$ $ \frac{n-1}{n}, \frac{n}{n}\}$, for an integer $n \geq 1$.

A {\em finitely many-valued interpretation} for $\alc$ is a pair $I=\langle \Delta, \cdot^I \rangle$ where:
$\Delta$ is a non-empty domain and 
$\cdot^I$ is an {\em interpretation function} that assigns 
to each $a \in N_I$ a value $a^I \in\Delta$, and
to each  $A\in N_C$ a function  $A^I :  \Delta \ri {\cal C}_n $
and to each role name $r \in N_R$  a function  $r^I:   \Delta \times  \Delta \ri {\cal C}_n $.
In particular, in \cite{ICLP22} we have considered two finitely many-valued fragments on $\alc$, 
the one based on \L ukasiewicz logic, and the other based on G\"odel logic extended with a standard involutive negation $ \ominus a = 1-a$.
Such fragments are defined along the lines of the finitely many-valued extension of description logic 
$\cal SROIQ$ \cite{BobilloStraccia_InfSci_11}, of the logic GZ $\cal SROIQ$ \cite{BobilloDelgadoStraccia2012},
and of the logic  
$\alc^*(S)$ \cite{GarciaCerdanaAE2010}. 

In the following, we will use ${\alc}_n$ to refer to a finitely-valued extension of $\alc$ interpreted over the truth space 
${\cal C}_n$, without committing to a specific choice of combination functions.
In Section \ref{sec:experimentation}, we will mainly refer to $G_n \lc$, the boolean fragment of ${\alc}_n$  based on G\"odel logic,
and we will assume that the interpretation of negated concepts exploits  involutive negation, i.e., $(\neg C)^I(x)= \ominus C^I(x) = 1-C^I(x)$.

\section{Fuzzy $\alc$ with typicality: $\alcFt$} \label{sec:fuzzyalc+T}

In this section, we extend fuzzy $\alc$ with typicality concepts of the form $\tip(C)$, where $C$ is a concept in fuzzy $\alc$.
The idea is similar to the extension of $\alc$ with typicality \cite{FI09},  
but transposed to the fuzzy case. The extension allows for the definition of {\em fuzzy typicality inclusions} of the form
$\tip(C) \sqsubseteq D \;\theta \;n$,   
meaning that typical $C$-elements are $D$-elements with a degree greater than $n$. A typicality  inclusion $\tip(C) \sqsubseteq D$, as in the two-valued case, stands for a KLM conditional implication $C \ent D$ \cite{KrausLehmannMagidor:90,whatdoes}, but now it has an associated degree.

We call $\alcFt$ the extension of fuzzy $\alc$ with typicality.
As in the two-valued case,  
such as in  $\sroiqpt$, a  preferential extension of ${\cal SROIQ}$ with typicality \cite{ISMIS2015},  or in the propositional typicality logic, PTL \cite{BoothCasiniAIJ19}, the typicality concept may be allowed to freely occur within inclusions and assertions, while the nesting of the typicality operator is not allowed.

In the definition of the semantics for $\alcFt$, we diverge from the choice in \cite{JELIA2021,ICLP22} and consider a fuzzy interpretation for the typicality operator, rather than a crisp one. This will allow us to prove that all the properties of a preferential consequence relation hold for a notion of entailment.

Observe that, in a fuzzy $\alc$ interpretation $I= \langle \Delta, \cdot^I \rangle$, the degree of membership $C^I(x)$ of the domain elements $x$ in a concept $C$ induces a preference relation $<_C$ on $\Delta$ as follows:
\begin{equation}\label{def:induced_order}
x <_C y \mbox{ iff } C^I(x) > C^I(y)
\end{equation}
Each preference  $<_{C}$ has the properties of preference relations in KLM-style ranked interpretations \cite{whatdoes}, that is,  $<_{C}$ is a modular and well-founded strict partial order.  
Let us recall that, $<_{C}$ is {\em well-founded} 
if there is no infinite descending chain $x_1 <_C x_0$, $x_2 <_C x_1$, $x_3 <_C x_2, \ldots $ of domain elements;
    $<_{C}$ is {\em modular} if,
for all $x,y,z \in \Delta$, $x <_{C} y$ implies ($x <_{C} z$ or $z <_{C} y$).
Well-foundedness holds for the induced preference $<_C$ defined by condition (\ref{def:induced_order}) under the assumption that  fuzzy interpretations are witnessed \cite{BobilloOWL2EL2018} (see Section \ref{sec:ALC}) or that $\Delta$ is finite. 

While each preference relation $<_C$ has  the properties of a preference relation in KLM  rational interpretations \cite{whatdoes} (also called ranked interpretations), here there are
multiple preferences and, therefore,
fuzzy interpretations can be regarded as {\em multipreferential} interpretations, which have also been studied in the two-valued case \cite{TPLP2020,Delgrande2020,AIJ21}. 

Each preference relation $<_C$ captures the relative typicality of domain elements wrt concept $C$ and may be used to identify the {\em typical  $C$-elements}. We will regard typical $C$-elements as the domain elements $x$ that  are preferred with respect to relation $<_C$
among those such that $C^I(x) \neq 0$.

For an interpretation $I$, let $C^I_{>0}$ be the crisp set containing all domain elements $x$ such that $C^I(x)>0$, that is, $C^I_{>0}= \{x \in \Delta \mid C^I(x)>0 \}$.
The (fuzzy) interpretation of typicality concepts $\tip(C)$ in $I$ is:
\begin{align}\label{eq:interpr_typicality}
	(\tip(C))^I(x)  & = \left\{\begin{array}{ll}
						 C^I(x) & \mbox{ \ \ \ \  if } x \in min_{<_C} (C^I_{>0}) \\
						0 &  \mbox{ \ \ \ \  otherwise } 
					\end{array}\right.
\end{align} 
where $min_{<_C}(S)= \{u: u \in S$ and $\nexists z \in S$ s.t. $z <_C u \}$.  When $(\tip(C))^I(x)>0$, we say that $x$ is a typical $C$-element in $I$.
Note that all typical $C$-elements have  the same membership degree in concept $C$.

 Observe also that, if $C^I(x)>0$ for some $x \in \Delta$,  
$min_{<_C} (C^I_{>0})$ is non-empty (and the size of the fuzzy concept $(\tip(C))^I$ is greater than zero).
This generalizes the property that, in the crisp case, $C^I\neq \emptyset$ implies  $(\tip(C))^I\neq \emptyset$.

Let us define a {\em fuzzy multi-preferential interpretation} for $\alcFt$ (shortly, an $\alcFt$ interpretation) as follows:

\begin{definition}[$\alcFt$ interpretation] \label{def:alcFt_interpretation}
\ An $\alcFt$ interpretation $I= \langle \Delta, \cdot^I \rangle$ is  fuzzy $\alc$ interpretation, equipped with the valuation of typicality concepts given by condition (\ref{eq:interpr_typicality}) above.
\end{definition}

The fuzzy interpretation  $I= \langle \Delta, \cdot^I \rangle$ implicitly defines a multi-preferential interpretation, where any concept $C$ is associated to a preference  relation $<_C$. 
This is different from 
the two-valued multi-preferential semantics in \cite{TPLP2020}, where only a subset of distinguished concepts have an associated preference, 
and a notion of global preference $<$ is introduced to define the interpretation of the typicality concept $\tip(C)$, for an arbitrary $C$. Here, we do not need to introduce a notion of global preference. The interpretation of any $\alc$ concept $C$ is defined compositionally from the interpretation of atomic concepts, and the preference relation $<_C$ associated to $C$ is defined from $C^I$.

The notions of {\em satisfiability} in $\alcFt$,   {\em model} of an $\alcFt$ knowledge base, and   $\alcFt$ {\em entailment} can be defined in a similar way as in fuzzy $\alc$ (see Section  \ref{sec:ALC}). 
In particular, given an $\alcFt$ knowledge base $K$,   
a fuzzy concept inclusion $\tip(C) \sqsubseteq D \; \theta \; k$ (with $\theta \in \{\geq, \leq, >, <\}$ and $k \in [0,1]$) is {\em entailed 
from $K$} in $\alcFt$ (written $K \models_{\alcFt} \tip(C) \sqsubseteq D \; \theta \; k$) 
if $\tip(C) \sqsubseteq D \; \theta k$ is satisfied in all $\alcFt$ models  
$I$ of the knowledge base $K$.  
In the following, we will refer to the entailment of $\tip(C) \sqsubseteq D \; \geq \; k$ as {\em $k$-entailment} and, as a special case, for $k=1$,  as {\em 1-entailment}.

As an example of satisfiability, the fuzzy concept inclusion $\langle \tip(C) \sqsubseteq D \geq k \rangle$  is satisfied in a fuzzy interpretation $I= \langle \Delta, \cdot^I \rangle$ if  $\inf_{x \in \Delta}  (\tip(C))^I(x) \rhd D^I(x) \geq k$ holds, 
which can be evaluated based on the combination functions of some specific fuzzy logic.

As in the two-valued case,  the typicality operator $\tip$ introduced in $\alcFt$  
is non-monotonic in the following sense: for a given knowledge base $K$, 
from the fact that $C \sqsubseteq D$ is 1-entailed from $K$, we cannot conclude that $ \tip(C) \sqsubseteq \tip(D)$  is 1-entailed from $K$.
Nevertheless, the logic $\alcFt$ is monotonic,
that is,  for two $\alcFt$ knowledge bases $K$ and $K'$, and a fuzzy axiom $E$, if $K \subseteq K'$, and $K \models_{\alcFt} E$ then 
$K' \models_{\alcFt} E$. 
 $\alcFt$ is a fuzzy relative of the monotonic logic $\alct$ \cite{FI09}.

Although, as we will see, the KLM postulates of a preferential consequence relation \cite{whatdoes} can be reformulated and hold for $\alcFt$, this typicality extension of fuzzy $\alc$ is rather weak. Similarly, in the two-valued case, the preferential extension of $\alc$ with typicality, 
$\alct$ \cite{FI09}, and the rational extension of $\alc$ with defeasible inclusions \cite{sudafricaniKR}
do not allow to deal with {\em irrelevance}.
From the fact that birds normally fly, one would like to be able to conclude that normally yellow birds fly, the color being irrelevant to flying. 

In the two-valued case, this has led to the definition of 
non-monotonic defeasible Description Logics \cite{casinistraccia2010,CasiniDL2013,AIJ15,BonattiSauro17,Casinistraccia2012,AIJ21},  which build on some closure construction (such as the rational closure   \cite{whatdoes} and the lexicographic closure \cite{Lehmann95} in  KLM framework) or some notion of minimal entailment \cite{bonattilutz}. 
In the next section, we introduce a notion of weighted knowledge base and strengthen $\alcFt$ by considering some different closure constructions, starting from the notion of coherent preferential interpretation introduced in \cite{JELIA2021}, and we discuss their properties.


\section{Weighted knowledge bases and closure constructions} \label{sec:closure}

To overcome the weakness of rational closure (as well as of preferential entailment), Lehmann introduced the lexicographic closure of a conditional knowledge base \cite{Lehmann95} which strengthens the rational closure by allowing further inferences.
From the semantic point of view, in the propositional case, a preference relation is defined on the set of propositional interpretations, so that the interpretations satisfying conditionals with higher rank are preferred to the interpretations satisfying conditionals with lower rank and, 
in case of contradictory defaults with the same rank,  interpretations satisfying more defaults with that rank are preferred.
The ranks of conditionals used by the  lexicographic closure construction are the ones computed by the rational closure construction \cite{whatdoes}, which capture  specificity:  the higher is the rank, the more specific is the default.
In other cases, the ranks may be part of the knowledge base specification, such as for ranked knowledge bases in Brewka's framework of basic preference descriptions  \cite{Brewka04}, or might be learned from empirical data, as we will see in the following.

In this section, we consider weighted (fuzzy) knowledge bases, where typicality inclusions are associated to weights, 
and develop a (semantic) closure construction to strengthen $\alcFt$ entailment,
which leads to some variants of the notion of fuzzy coherent multi-preferential model in \cite{JELIA2021}. 
The construction also relates to the definition of Kern-Isberner's c-representations  \cite{Kern-Isberner01,Kern-Isberner2014} which also include penalty points 
for falsified conditionals,
and to the algebraic semi-qualitative  approach to conditionals by Weydert \cite{Weydert03}.

A  {\em weighted $\alcFt$ knowledge base} $K$, over a set ${\cal C}= \{C_1, \ldots, C_k\}$ of distinguished $\alc$ concepts,
is a tuple $\langle  {\cal T}_{f}, {\cal T}_{C_1}, \ldots, {\cal T}_{C_k}, {\cal A}_f  \rangle$, where  ${\cal T}_{f}$  is a set of fuzzy $\alcFt$ inclusion axioms, ${\cal A}_f$ is a set of fuzzy $\alcFt$ assertions 
and, for each $C_i \in {\cal C}$,
${\cal T}_{C_i}=\{(d^i_h,w^i_h)\}$ is a (non-empty) set of weighted typicality inclusions $d^i_h= \tip(C_i) \sqsubseteq D_{i,h}$ for $C_i$, indexed by $h$, where each inclusion $d^i_h$ has weight $w^i_h$, a real number.
As in \cite{JELIA2021}, the typicality operator is assumed to occur only on the left hand side of a weighted typicality inclusion, and the {\em distinguished concepts} are those concepts $C_i$ occurring on the l.h.s.\ of some typicality inclusion $\tip(C_i) \sqsubseteq D$ in ${\cal T}_{C_i}$.
Arbitrary $\alcFt$ inclusions and assertions may belong to ${\cal T}_{f}$ and ${\cal A}_{f}$.

\begin{example} \label{exa:Penguin}
Consider the weighted knowledge base $K =\langle {\cal T}_{f},  {\cal T}_{Bird}, {\cal T}_{Penguin},$ $ {\cal T}_{Canary},$ $ {\cal A}_f \rangle$, over the set of distinguished concepts ${\cal C}=\{\mathit{Bird, Penguin,}$ 
$ \mathit{Canary}\}$, and assume the combination functions as in G\"odel fuzzy logic.
The Tbox $ {\cal T}_{f}$ contains the inclusions:

\begin{center}
$\mathit{Yellow \sqcap Black  \sqsubseteq  \bot} \geq 1$  
\ \   $\mathit{Yellow \sqcap Red  \sqsubseteq  \bot \geq 1}$   
\ \    $\mathit{Black \sqcap Red  \sqsubseteq  \bot \geq 1}$ 
\end{center}

\noindent
the ABox ${\cal A}_f$ contains the following assertions: 

\begin{quote} 
	$\mathit{Red(reddy)\geq 1 ,}$ $\mathit{\exists hasWings.Small(reddy)\geq 1,}$ 
	$\mathit{Fly(reddy) \geq 1}$
	 
	 $\mathit{Black(opus) \geq 1,}$ 
	 $\mathit{\exists hasWings.Long(opus)\geq 1}$, 
	 $\mathit{Fly(opus) \leq 0}$, 
\end{quote}

\noindent
the weighted TBox ${\cal T}_{Bird} $ 
contains the following weighted defeasible inclusions:

\smallskip
$(d_1)$ $\mathit{\tip(Bird) \sqsubseteq Fly}$, \ \  +20  \ \ \ \ \ \ \ \ \  \ \ \ \ \  

$(d_2)$ $\mathit{\tip(Bird) \sqsubseteq \exists hasWings. \top}$, \ \ +50

$(d_3)$ $\mathit{\tip(Bird) \sqsubseteq  \exists 
hasFeathering.\top}$, \ \ +50;

\smallskip
\noindent
and ${\cal T}_{Penguin}$ and  ${\cal T}_{Canary}$ contain, respectively, the following inclusions:

\smallskip
$(d_4)$ $\mathit{\tip(Penguin) \sqsubseteq Bird}$, \ \ +100 \ \ \ \ \ \ \ \   $(d_7)$ $\mathit{\tip(Canary) \sqsubseteq Bird}$, \ \ +100

$(d_5)$ $\mathit{\tip(Penguin) \sqsubseteq  Fly}$, \ \ - 70   \ \ \ \ \ \ \ \ \ \ \ \ \  $(d_8)$ $\mathit{\tip(Canary) \sqsubseteq Yellow}$, \ \  +30

$(d_6)$ $\mathit{\tip(Penguin) \sqsubseteq Black}$, \ \  +50; \ \ \ \ \ \ \   $(d_9)$ $\mathit{\tip(Canary) \sqsubseteq Red}$, \ \  +20

\smallskip
\noindent
 The intended meaning is that a bird normally has wings, has feathers and flies, but having wings and having feathers (both with weight 50)  for a bird is more plausible than flying (weight 20), although flying is regarded as being plausible. For a penguin, flying is not plausible (inclusion $(d_5)$ has negative weight -70), while being a bird and being black are very plausible properties of prototypical penguins, as $(d_4)$ and $(d_6)$ have positive weights (100 and 50, respectively). Similar considerations can be done for concept $\mathit{Canary}$. 
 
Consider an interpretation $I$, satisfying both TBox  and ABox axiams, in which Reddy is red, has small wings, has feathers and flies (suppose all with degree 1) and Opus has long wings, has feathers (with degree 1), is black with degree 0.8 and does not fly ($\mathit{Fly^I(opus^I) = 0}$). Considering the weights of defeasible inclusions, we might expect Reddy to be more typical than Opus as a bird, but less typical than Opus as a penguin in the interpretation $I$.

The fuzzy axioms in TBox ${\cal T}_f$ define strict constraints, e.g., for the second one, in any interpretation $I$, for each domain element $x$, the value of  $\mathit{(Red \sqcap Black)^I(x)}$ must be $0$  (and, hence, $\mathit{Black^I(reddy^I) = 0}$).
Note also that, as ABox ${\cal A}_f$ contains the assertion  $\mathit{\exists hasWings.Small(reddy)\geq 1}$, 
then  $\mathit{(\exists hasWings.Small)^I (reddy^I)}$ $= 1$ holds.
Hence, there is a domain element $y \in \Delta$ such that $\mathit{ hasWings^I(reddy^I,}$ $\mathit{  y)=1}$ and $\mathit{ Small^I(y)=1}$.
Thus, it follows that $\mathit{(\exists hasWings.\top)^I (reddy^I)= 1}$,
and hence the assertion $\mathit{\exists hasWings.\top}$ $\mathit{ (reddy)\geq 1}$ is as well satisfied in $I$.
 \end{example}

We define the semantics of weighted knowledge bases as the one above through a {\em semantic closure construction}, similar in spirit to Lehmann's lexicographic closure \cite{Lehmann95}, but exploiting weights and based on multiple preferences.
The construction allows a subset of the $\alcFt$ interpretations to be selected, 
the interpretations whose induced preference relations $<_{C_i}$, for the distinguished concepts $C_i$,  faithfully represent the defeasible part of the knowledge base $K$.

Let ${\cal T}_{C_i}=\{(d^i_h,w^i_h)\}$ be the set of weighted typicality inclusions $d^i_h= \tip(C_i) \sqsubseteq D_{i,h}$ associated to the distinguished concept $C_i$, and let $I=\langle \Delta, \cdot^I \rangle$ be a fuzzy $\alcFt$ interpretation.
In the two-valued case, we would associate to each domain element $x \in \Delta$ and each distinguished concept $C_i$, a weight $W_i(x)$ of $x$ wrt $C_i$ in $I$, by summing the weights of the defeasible inclusions satisfied by $x$.
However, as $I$ is a fuzzy interpretation, we do not only  distinguish between the typicality inclusions satisfied or 
falsified  by $x$;
 we also need to consider, for all inclusions $\tip(C_i) \sqsubseteq D_{i,h} \in {\cal T}_{C_i}$,  
the degree of membership of $x$ in $D_{i,h}$. 
Furthermore, in comparing the weight of domain elements with respect to $<_{C_i}$, we want to give higher preference to the domain elements having a membership degree in $C_i$ greater than $0$, 
with respect to those elements whose degree of membership in $C_i$ is $0$.

For each domain element $x \in \Delta$ and distinguished concept $C_i$, {\em the weight $W_i(x)$ of $x$ wrt $C_i$} in the $\alcFt$ interpretation $I=\langle \Delta, \cdot^I \rangle$ is defined as follows:
 \begin{align}\label{weight_fuzzy}
	W_i(x)  & = \left\{\begin{array}{ll}
						 \sum_{h} w_h^i  \; D_{i,h}^I(x) & \mbox{ \ \ \ \  if } C_i^I(x)>0 \\
						- \infty &  \mbox{ \ \ \ \  otherwise }  
					\end{array}\right.
\end{align} 
where $-\infty$ is added at the bottom of all real values.

The value of $W_i(x) $ is $- \infty $ when $x$ is not a $C$-element (i.e., $C_i^I(x)=0$). 
Otherwise, $C_i^I(x) >0$ and the higher is the sum $W_i(x) $, the more typical is the element $x$ relative to concept $C_i$.
How much $x$ satisfies a typicality property  $\tip(C_i) \sqsubseteq D_{i,h}$ depends on the value of $D_{i,h}^I(x) \in [0,1]$, which is weighted by $ w_h^i $ in the sum. 
In the two-valued case, $D_{i,h}^I(x) \in \{0,1\}$, and 
$W_i(x)$ is the sum of the weights of the typicality inclusions for $C$ satisfied by $x$, if $x$ is a $C$-element,  and is $-\infty $, otherwise.

\begin{example} \label{exa:penguin2}
Let us continue Example \ref{exa:Penguin}.
In the $\alcFt$ interpretation  $I$, it holds that: 
$\mathit{Fly^I(reddy^I)  = (\exists has\_Wings. \top)^I (reddy^I)}$ = $\mathit{(\exists has\_Feathering. }$ $\mathit{\top)^I (reddy^I)=}$ 
 $ \mathit{Red^I(reddy^I) =1 }$,
i.e., Reddy   flies, has wings and feathers and is red (and hence $\mathit{Black^I(reddy^I)=0}$). 
For Opus it holds that: 
$\mathit{Fly^I(opus^I) =}$ $\mathit{ 0}$,
$\mathit{ Black^I(opus^I) =}$   $\mathit{ 0.8}$ and
$\mathit{ (\exists has\_Wings. \top)^I (opus^I)}$
$ \mathit{= (\exists has\_Feathering. }$ $\mathit{\top)^I (opus^I)=1 }$, i.e., Opus does not fly, is black with degree 0.8, it has wings and feathers.

Let us further assume that $\mathit{Bird^I(reddy^I)=1}$ and $\mathit{Bird^I(opus^I)}$ $=0.8$.
Considering the weights of the typicality inclusions for $\mathit{Bird}$:  

\smallskip
$\mathit{W_{Bird}(reddy^I)= 20+50+50=120}$ 

$\mathit{W_{Bird}(opus^I)= 0+50+50=100}$

\smallskip
\noindent
which suggests that Reddy should be more typical as a bird than Opus.

On the other hand, if we suppose $\mathit{Penguin^I(reddy^I)=0.2}$ and $\mathit{Penguin^I}$ $\mathit{(opus^I)=}$ $0.8$,
we have: 

\smallskip
$\mathit{W_{Penguin}(reddy)}$ $ \mathit{= 100-70+0=30}$ 

$\mathit{W_{Penguin}(opus)= 0.8 \times 100} $ $\mathit{ -0+0.8 \times 50=120}$.

\smallskip
\noindent
This suggests that Reddy should be less typical as a penguin than Opus.
 \end{example}
We have seen in Section \ref{sec:fuzzyalc+T} that each fuzzy interpretation $I$ induces a preference relation for each concept and, in particular, it induces a preference   $<_{C_i}$ for each distinguished concept $C_i$. 
We further require that, if $x  <_{C_i}  y$, 
then $x$ must be more typical than $y$ wrt $C_i$, that is, 
 the weight $W_i(x)$ of $x$ wrt $C_i$ should be higher than the weight $W_i(y)$ of $y$ wrt $C_i$ (and $x$ should satisfy more properties or more plausible properties of typical $C_i$-elements with respect to $y$). 
This leads to the following definition of {\em faithful multi-preferential model} of a weighted a $\alcFt$  knowledge base.

\begin{definition}[Faithful (fuzzy) multi-preferential model of $K$]\label{fuzzy_fm-model} 
Let $K=\langle  {\cal T}_{f},$ $ {\cal T}_{C_1}, \ldots,$ $ {\cal T}_{C_k}, {\cal A}_f  \rangle$ be  a weighted $\alcFt$ knowledge base  over  ${\cal C }$.  
A {\em  faithful (fuzzy) multi-preferential model} (fm-model)  of $K$ is  a fuzzy $\alcFt$ interpretation $I=\langle \Delta, \cdot^I \rangle$ 
s.t.: 
\begin{itemize}
\item
$I$  satisfies  the fuzzy inclusions in $ {\cal T}_{f}$ and the fuzzy assertions in ${\cal A}_f$;
\item  
for all $C_i\in {\cal C}$,  the preference {\em $<_{C_i}$   is faithful to $ {\cal T}_{C_i}$}, that is: 
\begin{align}\label{weak_agreement} 
x  <_{C_i}  y & \Ri W_i(x) > W_i(y)  
\end{align}
\end{itemize}
\end{definition}

\begin{example}
Referring to Example  \ref{exa:penguin2} above, clearly, $reddy <_{Bird} opus$, 
as $\mathit{Bird^I(reddy)=1}$ and $\mathit{Bird^I(opus)}$ $=0.8$,  while $opus <_{Penguin} reddy$, as $\mathit{Penguin^I(reddy)=0.2}$ and $\mathit{Penguin^I(opus)=}$ $0.8$.
For the interpretation $I$ to be faithful, it is necessary that the conditions $\mathit{W_{Bird}(reddy) > W_{Bird}(opus)}$ and $\mathit{W_{Penguin}}$ $\mathit{(opus) >}$ $\mathit{ W_{Penguin}(reddy)}$ hold with respect to interpretation $I$. This is true, as we have seen in Example  \ref{exa:penguin2}.
On the contrary, if we had $\mathit{Penguin^I(reddy)=0.9}$, the interpretation $I$ would not be faithful (as it assigns to $\mathit{reddy}$ a membership degree in concept $\mathit{Penguin}$ higher than the one for $\mathit{opus}$).
\end{example}

Let us now consider two alternative closure constructions, by introducing the notions of {\em coherent } and of {\em $\varphi$-coherent} models.

The notion of {\em coherent (fuzzy) multi-preferential model} of $K$, can be defined as in Definition \ref{fuzzy_fm-model} above, but replacing the faithfulness condition (\ref{weak_agreement}), with the following stronger coherence condition:
\begin{align}\label{pref_rel_fuzzy}
x  <_{C_i}  y & \mbox{  \ \ iff \ \ } W_i(x) > W_i(y)  
\end{align}
This is a reformulation of the notion of coherent (fuzzy) multi-preferential model from \cite{JELIA2021}, but here we do not restrict to a crisp interpretation of typicality concepts $\tip(C)$.

The weaker notion of faithfulness determines a larger class of fuzzy multi-preferential models of a weighted knowledge base, compared to the class of coherent models. 
As we will see in Section  \ref{sec:multiulayer_perceptron}, this also allows a larger class of monotone non-decreasing activation functions in neural network models to be captured.

The notion of {\em $\varphi$-coherence} of a fuzzy interpretation $I$ wrt a KB, first introduced in  \cite{WorkshopAI3}, exploits a function $\varphi$  from $\mathbb{R}$ to the interval $[0,1]$, i.e., $\varphi: {\mathbb{R}} \rightarrow [0,1]$.
We actually allow for possibly different functions $\varphi_i: {\mathbb{R}} \rightarrow [0,1]$, one for each concept $C_i \in {\cal C}$.
As we will see, $\varphi$ or the $\varphi_i$ are intended to represent the activation function(s) for units in a neural network $\enne$.
 
 \begin{definition}[$\varphi$-coherence]\label{varphi-coherence} 
 Let $K=\langle  {\cal T}_{f},$ $ {\cal T}_{C_1}, \ldots,$ $ {\cal T}_{C_k}, {\cal A}_f  \rangle$ be  a weighted $\alcFt$ knowledge base, and $\varphi$ a collection of functions $\varphi_i: {\mathbb{R}} \rightarrow [0,1]$, for $i= 1, \ldots, k$.
A fuzzy $\alcFt$ interpretation $I=\langle \Delta, \cdot^I \rangle$  is {\em $\varphi$-coherent} if, 
for all concepts $C_i \in {\cal C}$ and $x\in \Delta$, 
\begin{align}
\label{fi_coherence}
C_i^I(x)= \varphi_i (\sum_{h} w_h^i  \; D_{i,h}^I(x)) 
\end{align}
where ${\cal T}_{C_i}=\{(\tip(C_i) \sqsubseteq D_{i,h},w^i_h)\}$ is the set of weighted conditionals  for $C_i$.
\end{definition}
Observe that,  for all $x$ such that $C_i(x)>0$, condition (\ref{fi_coherence}) above corresponds to condition $C_i^I(x)= \varphi_i (W_i(x))$, for all distinguished concepts $C_i \in {\cal C}$.
While in coherent and faithful models the notion of weight $W_i(x)$ 
considers, as a special case, the case $C_i(x)=0$, condition (\ref{fi_coherence}) imposes the same constraint to all domain elements $x$. 

{\em Coherent} (resp., {\em $\varphi$-coherent}) {\em multi-preferential models} of a knowledge base $K$, can be defined similarly to faithful models in Definition \ref{fuzzy_fm-model}.
We provide explicitly the definition of $\varphi$-coherent model of $K$.
\begin{definition}[$\varphi$-coherent (fuzzy) multi-preferential  model of $K$]\label{def:varphi-model} 
Let $K=\langle  {\cal T}_{f},$ $ {\cal T}_{C_1}, \ldots,$ $ {\cal T}_{C_k}, {\cal A}_f  \rangle$ be  a weighted $\alcFt$ knowledge base  over  ${\cal C }$.  
A {\em  $\varphi$-coherent (fuzzy) multi-preferential model} (or, simply, $\varphi$-coherent model)  of $K$ is  a fuzzy $\alcFt$ interpretation $I=\langle \Delta, \cdot^I \rangle$ 
s.t.: 
\begin{itemize}
\item
$I$  satisfies  the fuzzy inclusions in $ {\cal T}_{f}$ and the fuzzy assertions in ${\cal A}_f$;
\item for all the distinguished concepts $C_i \in {\cal C}$, for all $x \in \Delta$,
\begin{align*}
C_i^I(x)= \varphi_i (\sum_{h} w_h^i  \; D_{i,h}^I(x)) 
\end{align*}
where ${\cal T}_{C_i}=\{(\tip(C_i) \sqsubseteq D_{i,h},w^i_h)\}$ is the set of all the weighted conditionals  for $C_i$.

\end{itemize}
\end{definition}


The following proposition establishes the relationships between  $\varphi$-coherent, faithful and coherent  fuzzy multi-preferential models of  a weighted conditional knowledge base $K$.

\begin{proposition}   \label{prop:phi_coherent_models}
Let $K$ be a weighted conditional $\alcFt$ knowledge base and let $\varphi_i: {\mathbb{R}} \rightarrow [0,1]$, for all $i=1, \ldots, k$.
The following statements hold:
\begin{itemize}
\item[(1)] Any coherent model of $K$ is a  faithful model of $K$;

\item[(2)] If the $\varphi_i$ are {\em monotonically non-decreasing} functions, a $\varphi$-coherent multi-preferential model $I$ of $K$ is also a faithful model of $K$;

\item[(3)] If the $\varphi_i$ are {\em monotonically increasing} functions, a $\varphi$-coherent multi-preferential model $I$ of $K$ is also a coherent-model of $K$. 
\end{itemize}
\end{proposition}

\begin{proof} 
Item (1) directly follows from the definition, as the coherence condition (\ref{pref_rel_fuzzy}) is stronger than the faithfulness condition (\ref{weak_agreement}).

For item (2),  let us assume that the $\varphi_i$ are monotonically non-decreasing functions and that $I=\langle \Delta, \cdot^I \rangle$ is a $\varphi$-coherent fuzzy multi-preferential model of $K$. In particular, for all distinguished concepts $C_i$ and $z\in \Delta$, s.t. $C_i^I(z)>0$, $C_i^I(z)= \varphi_i (W_i(z))$. 
To prove condition (\ref{weak_agreement}), i.e., that $x  <_{C_i}  y \Ri W_i(x) > W_i(y)$ holds, let us assume that, for some $x,y \in \Delta$, $x  <_{C_i}  y$ holds, i.e., $C_i^I(x) > C_i^I(y)$. This also implies $C_i^I(x)>0$.
If $C_i^I(y)=0$, $W_i(y)= - \infty$, and the thesis follows.
If $C_i^I(y)>0$,  both the equalities $C_i^I(x)= \varphi_i (W_i(x))$ and $C_i^I(y)= \varphi_i (W_i(y))$ hold.
 Suppose that $W_i(x) > W_i(y)$ does not hold, i.e.,  that $W_i(x) \leq W_i(y)$.
As $\varphi_i$ is monotonically non-decreasing, $\varphi_i(W_i(x)) \leq \varphi_i (W_i(y))$.
Hence, by the equalities above, $C_i^I(x) \leq C_i^I(y) $, 
which contradicts the assumption that $C_i^I(x) > C_i^I(y)$.

For item (3), assume that the $\varphi_i$ are monotonically increasing functions and that $I=\langle \Delta, \cdot^I \rangle$ is a $\varphi$-coherent multi-preferential model of $K$. Then, equality (\ref{fi_coherence}) holds.
In particular, for all distinguished concept $C_i$ and $z\in \Delta$, s.t. $C_i^I(z)>0$, $C_i^I(z)= \varphi_i (W_i(z))$. 

We have to prove that condition (\ref{pref_rel_fuzzy}) 
holds, i.e., that
$x  <_{C_i}  y$ iff $ W_i(x) > W_i(y)$. The ``only if" direction holds with the same proof as for item (2), as $\varphi_i$ is as well monotonically non-decreasing.
To prove the ``if" direction, assume that  $ W_i(x) > W_i(y)$ holds, for some $x,y \in \Delta$. 
If  $W_i(y)= - \infty$, it must be that $C_i^I(y)=0$ and $C_i^I(x)>0$, and hence $C^I(x) > C^I(y)$ follows.
If $W_i(y)\neq - \infty$, $C_i^I(y)>0$ and $C_i^I(x)>0$.
From $ W_i(x) > W_i(y)$, as $\varphi_i$ is monotonically increasing,
$ \varphi_i (W_i(x) ) > \varphi_i(W_i(y))$. Hence, $C_i^I(x) > C_i^I(y)$.  
\end{proof}

The notions of {\em faithful/coherent/$\varphi$-coherent multi-preferential entailment} 
from a weighted $\alcFt$ knowledge base $K$ can be defined 
as expected.
\begin{definition}[Faithful/coherent/$\varphi$-coherent entailment] \label{fm-entailment}
A fuzzy axiom $E$   is {\em faithfully entailed}  (resp., {\em coherently/$\varphi$-coherently entailed})
from a fuzzy weighted knowledge base $K$ (for short $K \models_{fm/cm/\varphi} E$)  if, for all faithful models (resp., coherent/$\varphi$-coherent-models) $I=\langle \Delta, \cdot^I \rangle$ of $K$, $I$ satisfies $E$. 
\end{definition}

As usual in preferential semantics, a stronger notion of entailment can be obtained by restricting to a specific subset of models, namely, to {\em canonical models}, which are large enough to contain all the relevant domain elements. More precisely, in the two-valued case, a canonical model contains
a domain element for each possible valuation of concepts which is present in some model of K  \cite{AIJ15,TPLP2020}.
The notion of canonical model can be extended to the many-valued case.

\begin{definition}\label{def-canonical-model-DL} 
Given a weighted knowledge base $K=\langle  {\cal T}_f, {\cal T}_{C_1}, \ldots, {\cal T}_{C_k}, {\cal A}_f  \rangle$ a faithful/coherent/$\varphi$-coherent  $\alcFt$ model $I=\langle \Delta, \cdot^{\cal I} \rangle$ of $K$ is {\em canonical} if, for each faithful/coherent/$\varphi$-coherent model $I'= \langle \Delta',  \cdot^{\cal '} \rangle$ of $K$, and for each $x \in \Delta'$, there is an element $y \in \Delta$ such that $A^I(y)=A^{I'}(x)$, for all concept names $A$ occurring in $K$.
\end{definition}
That is,  a canonical faithful/coherent/$\varphi$-coherent model for $K$ contains a domain element $y$ corresponding to each domain element $x$ in any faithful/ coherent/$\varphi$-coherent interpretation $I'$ of $K$, and $y$ has the same membership degree as $x$ in all named concepts $A$ occurring in $K$. 
Note that, as in the two-valued case for defeasible $\alc$ \cite{Casini21} (and similarly for $\alc$ with typicality \cite{AIJ15}), also in the many-valued case two $\alcFt$ models of the knowledge base can be combined by taking the disjoint union of their domains, to construct a new (larger) model of the KB (and the proof is similar to the one in the two-valued case \cite{Casini21}). This property guarantees that, if a faithful/coherent/$\varphi$-coherent  $\alcFt$ model of a knowledge base $K$ exists, a canonical faithful/coherent/$\varphi$-coherent  $\alcFt$ model of $K$ also exists.

A notion of {\em canonical faithful/coherent/$\varphi$-coherent entailment} can be defined. In the following, we assume that axiom $E$ only contains concept names occurring in the knowledge base $K$.

\begin{definition}[Canonical entailment] \label{canonical-entailment}
Given a weighted $\alcFt$  knowledge base $K$, 
a  fuzzy axiom $E$  is {\em canonically entailed} from $K$ 
in the faithful/ coherent/$\varphi$-coherent  semantics if, 
for all  {\em canonical faithful/coherent/}
\linebreak
{\em $\varphi$-coherent models} $I=\langle \Delta, \cdot^I \rangle$ of $K$, $I$ satisfies $E$.
\end{definition}

In Sections \ref{sec:KLM} and   \ref{sec:multiulayer_perceptron}  we study the KLM properties of $\alcFt$ and its relationships with MLPs, under the different closure constructions.

\section{The KLM properties of $\alcFt$ and its closures} \label{sec:KLM}

In this section we investigate whether  the KLM postulates of a preferential consequence relation \cite{KrausLehmannMagidor:90,whatdoes} are satisfied by  entailment in $\alcFt$ as well as in the coherent and faithful semantics. 

The satisfiability of KLM postulates of rational or preferential consequence relations \cite{KrausLehmannMagidor:90,whatdoes} has been studied for $\alc$ with defeasible inclusions and typicality inclusions in the two-valued case \cite{sudafricaniKR,FI09,BritzTOCL2021}. The KLM postulates of a preferential consequence relation (namely, Reflexivity, Left Logical Equivalence, Right Weakening, And, Or, Cautious Monotonicity)
can be reformulated for $\alc$ with typicality, by considering that a typicality inclusion $\tip(C) \sqsubseteq D$ stands for a conditional $C {\ent} D$ in KLM preferential logics, by the following properties, expressed as inference rules:

\begin{quote}

$(REFL)$ \ $\tip(C) \sqsubseteq C $

$(LLE)$ \ If $\models A \equiv B$ and $\tip(A) \sqsubseteq C $, then $\tip(B) \sqsubseteq C $ 

$(RW)$ \  If $\models C \sqsubseteq D$ and $\tip(A) \sqsubseteq C $, then $\tip(A) \sqsubseteq D $ 

$(AND)$ \ If $\tip(A) \sqsubseteq C $ and $\tip(A) \sqsubseteq D $, then $\tip(A) \sqsubseteq C \sqcap D $ 

$(OR)$ \ If $\tip(A) \sqsubseteq C $ and $\tip(B) \sqsubseteq C $, then $\tip(A \sqcup B) \sqsubseteq C $

$(CM)$ \  If $\tip(A) \sqsubseteq D$ and $\tip(A) \sqsubseteq C $, then $\tip(A \sqcap D) \sqsubseteq C $

\end{quote}
where $\models A \equiv B$ is interpreted as equivalence of concepts $A$ and $B$ in the underlying description logic $\alc$ (i.e., $A^I = B^I$ in all $\alc$ interpretations $I$), while $\models C \sqsubseteq D$
is interpreted as validity of the inclusion $C \sqsubseteq D$ in $\alc$ (i.e.,  $A^I \subseteq B^I$ for all $\alc$ interpretations $I$).

The interpretation of the postulates is that, given a knowledge base $K$ in $\alc$ extended with typicality inclusions, the logical consequences of $K$ satisfy the properties above. For instance,  reflexivity $(REFL)$ requires that $\tip(C) \sqsubseteq C $  is a logical consequence of $K$; for Left Logical Equivalence $(LLE)$, if $A$ and $B$ are logically equivalent in $\alc$ and
$\tip(A) \sqsubseteq C $ is a logical consequence of $K$ , then $\tip(B) \sqsubseteq C $ must be as well a logical consequence of $K$; and so on.

In the following we reformulate these postulates for the fuzzy case and, specifically for $\alcFt$.
We reinterpret $\models C \sqsubseteq D$ as the requirement that the fuzzy inclusion $C \sqsubseteq D \geq 1$ is valid in fuzzy $\alc$ (that is, $C \sqsubseteq D \geq 1$ is satisfied in all fuzzy $\alc$ interpretations),
and $\models A\equiv B$ as the requirement that the fuzzy inclusions $A \sqsubseteq B \geq 1$ and $B \sqsubseteq A \geq 1$ are valid in fuzzy $\alc$.
Interpreting inclusions of the form  $\tip(A) \sqsubseteq C $ as fuzzy inclusions $\tip(A) \sqsubseteq C \geq 1$, we reformulate the KLM postulates for $1$-entailment:

\begin{quote}
$\mathit{(REFL')}$ \ $\tip(C) \sqsubseteq C \geq 1$ 

$\mathit{(LLE')}$ \ If 
$\models A \equiv B$  and $\tip(A) \sqsubseteq C \geq 1$, then $\tip(B) \sqsubseteq C  \geq 1$ 

$\mathit{(RW')}$ \  If $\models C \sqsubseteq D$ and $\tip(A) \sqsubseteq C \geq 1$,
then $\tip(A) \sqsubseteq D \geq 1$ 

$\mathit{(AND')}$ \ If $\tip(A) \sqsubseteq C \geq 1 $ and $\tip(A) \sqsubseteq D \geq 1$, \\
$\mbox{ }$ \ \ \ \ \ \ \ \ \ then $\tip(A) \sqsubseteq C \sqcap D \geq 1$

$\mathit{(OR')}$ \ If $\tip(A) \sqsubseteq C \geq 1$ and $\tip(B) \sqsubseteq C \geq 1$, \\
$\mbox{ }$ \ \ \ \ \ \ \ \ \ then $\tip(A \sqcup B) \sqsubseteq C \geq 1$

$\mathit{(CM')}$ \  If $\tip(A) \sqsubseteq D \geq 1$ and $\tip(A) \sqsubseteq C \geq 1$, \\
$\mbox{ }$ \ \ \ \ \ \ \ \ \ then $\tip(A \sqcap D) \sqsubseteq C \geq 1$ 
\end{quote}
As an example, the meaning of right weakening $\mathit{(RW')}$ is that, if it holds that  $\models C \sqsubseteq D$ (i.e., $C \sqsubseteq D \geq 1$ is valid in fuzzy $\alc$),
and  $\tip(A) \sqsubseteq C \geq 1$ is entailed from a weighted knowledge base $K$, then $\tip(A) \sqsubseteq D \geq 1$ is also entailed by $K$.

To prove that all the postulates above hold for the choice of combination functions as in G\"odel logic, we prove that each postulate is satisfied in any $\alcFt$ interpretation.
For instance, for $\mathit{(RW')}$ this means that, if it holds that  $\models C \sqsubseteq D$ (i.e., $C \sqsubseteq D \geq 1$ is valid in fuzzy $\alc$), then in any $\alcFt$ interpretation $I$, if $\tip(A) \sqsubseteq C \geq 1$ is satisfied in $I$, then $\tip(A) \sqsubseteq D \geq 1$ is also satisfied in $I$.

\begin{proposition} \label{prop:KLM_properties}
Under the choice of combination functions as in G\"odel logic, any $\alcFt$ interpretation $I=\langle \Delta, \cdot^I \rangle $
satisfies the postulates $\mathit{(REFL')}$, $\mathit{(LLE')}$, $\mathit{(RW')}$, $\mathit{(AND')}$, $\mathit{(OR')}$ and $\mathit{(CM')}$.
\end{proposition}

\noindent
The proof of Proposition \ref{prop:KLM_properties} can be found in the Appendix. 
As a simple consequence, the following corollary states that, for the choice of combination functions as in G\"odel logic, $1$-entailment in $\alcFt$ 
satisfies the KLM postulates $\mathit{(REFL')}$, $\mathit{(LLE')}$, $\mathit{(RW')}$, $\mathit{(AND')}$, $\mathit{(OR')}$ and $\mathit{(CM')}$.

\begin{corollary} \label{corollary:KLM}
For the choice of combination functions as in G\"odel logic, $1$-entailment in $\alcFt$ 
satisfies the KLM postulates $\mathit{(REFL')}$, $\mathit{(LLE')}$, $\mathit{(RW')}$, $\mathit{(AND')}$, $\mathit{(OR')}$ and $\mathit{(CM')}$.
\end{corollary}

\begin{proof}[Proof (Sketch)]
Consider, for instance, postulate $\mathit{(AND')}$. Assume $\tip(A) \sqsubseteq C \geq 1 $ and $\tip(A) \sqsubseteq D \geq 1$ are entailed from a knowledge base $K$ in $\alcFt$. Then they are satisfied in all $\alcFt$ models $I$ of $K$. Hence,  by Proposition  \ref{prop:KLM_properties}, $\tip(A) \sqsubseteq C \sqcap D \geq 1$ is also satisfied in all the models $I$ of $K$, i.e., $\tip(A) \sqsubseteq C \sqcap D \geq 1$ is entailed by $K$.
The proof of all other properties is similar.
\end{proof}

\noindent
Corollary  \ref{corollary:KLM} tells us that, for the choice of combination functions as in G\"odel logic, 
$1$-entailment in $\alcFt$ 
satisfies the properties 
of a preferential consequence relation. 
Observe that this result does not depend on the choice of the negation combination function as negation does not occur in the postulates we have considered; in particular, the result holds as well for G\"odel logic with standard involutive negation. 
On the other hand, 
1-entailment in $\alcFt$ does not satisfy the Rational Monotonicity postulate, so it does not satisfy all postulates  of a rational consequence relation. 
Let us reformulate the property of Rational Monotonicity in the fuzzy case as follows:

\noindent
$\mathit{(RM')}$ \  If $\tip(A) \sqsubseteq C \geq 1$ and not $\tip(A) \sqsubseteq \neg B \geq 1$, then $\tip(A \sqcap B) \sqsubseteq C \geq 1$ 

\begin{proposition}\label{prop:RM}
For the choice of combination functions as in G\"odel logic, $\mathit{(RM')}$ does not hold in $\alcFt$
(and the same for G\"odel logic with standard involutive negation).
\end{proposition}
The proof of the proposition in the Appendix  provides a counterexample to Rational Monotonicity for a knowledge base without weighted inclusions.

The postulates for $1$-entailment considered above may be violated by other choices of combination functions.
For instance, the choice of combination functions as in Product logic or as in \L ukasiewicz logics fails to satisfy both postulates $\mathit{(AND')}$ and $\mathit{(OR')}$.
The postulates  $\mathit{(REFL')}$, $\mathit{(LLE')}$, $\mathit{(RW')}$, $\mathit{(AND')}$, $\mathit{(OR')}$ and $\mathit{(CM')}$ can as well be formulated for  {\em $k$-entailment}, by replacing the occurrences of typicality inclusions $\tip(A) \sqsubseteq C \geq 1$ with $\tip(A) \sqsubseteq C \geq k$.
For combination functions as in G\"odel logic, all the postulates for $k$-entailment hold (with a proof similar to the one for 1-entailment), except for Cautious Monotonicity $\mathit{(CM)}$, which does not hold.

For faithful, coherent and $\varphi$-coherent entailment,
 the next corollary also follows from Proposition \ref{prop:KLM_properties} as a simple consequence, by observing that all faithful, coherent and $\varphi$-coherent models of a knowledge base $K$ are $\alcFt$ models of $K$. 

\begin{corollary}
For the choice of combination functions as in G\"odel logic, faithful, coherent and $\varphi$-coherent entailment in $\alcFt$ 
satisfy postulates $\mathit{(REFL')}$, $\mathit{(LLE')}$, $\mathit{(RW')}$, $\mathit{(AND')}$, $\mathit{(OR')}$ and $\mathit{(CM')}$ of 1-entailment.
\end{corollary}

The results above improve over the previous results in \cite{ECSQARU2021}, which have been proven for a crisp interpretation of the typicality concept.
When the interpretation of $\tip(C)$ is either $0$ or $1$, 1-entailment in $\alcFt$ fails to satisfy the Reflexivity postulate $(REFL')$.

Some further properties of typicality can be obtained by reformulating for the fuzzy case  the semantic properties of $\alc+T$ in \cite{FI09}. We name the properties $(f_T - 1), \ldots, (f_T - 5)$ after \cite{FI09}:
\begin{quote}
$(f_T - 1)$ \ $\tip(C) \sqsubseteq C \geq 1$      

$(f_T - 2)$ if $\tip(C) \equiv \bot$, then $C \equiv \bot$

$(f_T - 3)$
If  $\tip(A) \sqsubseteq D \geq 1$, then $\tip(A) \equiv  \tip(A \sqcap D)$ 

$(f_T - 4)$
 $\tip(A \sqcup B) \sqsubseteq \tip(A) \sqcup \tip(B) \geq 1$

$(f_T - 5)$
 $\tip(A) \sqcap \tip(B) \sqsubseteq  \tip(A \sqcup B)  \geq 1$.

\end{quote}
where, for two $\alcFt$ concepts, $C \equiv D$, stands for $(C \sqsubseteq D \geq 1) \sqcap (D \sqsubseteq C \geq 1)$.

Note that $(f_T - 1)$ is $\mathit{(REFL')}$; $(f_T - 2)$ is a consequence of well-foundedness of the preference relations;
$(f_T - 3)$ implies $\mathit{(CM)}$; and  $(f_T - 4)$ is a reformulation of $\mathit{(OR)}$.
It can be proven that properties $(f_T - 1), \ldots, (f_T - 5)$ are satisfied in all $\alcFt$ interpretations. 
\begin{proposition} \label{prop:typicality_properties}
Under the choice of combination functions as in G\"odel logic, any $\alcFt$ interpretation satisfies the postulates $(f_T - 1), \ldots, (f_T - 5)$.
\end{proposition} 
The proof is similar to the proof of Proposition  \ref{prop:KLM_properties}.
To conclude this section let us 
informally describe how fuzzy multi-preferential entailment deals with irrelevance and avoids inheritance blocking, properties which have been considered  as desiderata for preferential logics of defeasible reasoning \cite{Weydert03,Kern-Isberner2014}.

Concerning ``irrelevance", let us consider again previous Example  \ref{exa:Penguin}: if typical birds fly, we would like to conclude that  typical yellow birds also fly, as the property of being yellow is irrelevant with respect to flying. Observe, that in Example  \ref{exa:penguin2}, we can conclude that Reddy is more typical than Opus  as a bird ($\mathit{reddy <_{Bird} opus}$), as Opus does not fly, while Reddy flies. The relative typicality of Reddy and Opus wrt $\mathit{Bird}$ does not depend on their color (the weighted TBox ${\cal T}_{Bird} $  does not refer to a color) and we would obtain the same relative preferences if Reddy were yellow rather than red. 

The  fuzzy multi-preferential entailment is not subject to the problem called by Pearl  the ``blockage of property inheritance" problem  \cite{Pearl90},
and by Benferhat et al. the ``drowning problem"  \cite{BenferhatIJCAI93}. This problem affects the rational closure and system Z \cite{Pearl90}, as well as the rational closure refinements. Roughly speaking, the problem is that property inheritance from classes to subclasses is not guaranteed.
If a subclass is exceptional with respect to a superclass for a given property, 
it does not inherit from that superclass any other property.  
For instance, referring to the typicality inclusions in Example \ref{exa:penguin2}, in the rational closure, typical penguins would not inherit the property of typical birds of having wings, being exceptional to birds concerning flying.
On the contrary,
in  fuzzy multi-preferential models, 
considering again Example \ref{exa:penguin2}, the degree of membership of a domain element $x$ in concept $\mathit{Bird}$, i.e., $\mathit{Bird^I(x)}$, is used to determine the weight of $x$ with respect to $\mathit{Penguin}$. As the weight of typicality inclusion $(d_4)$ is positive, the higher is the value of $\mathit{Bird^I(x)}$, the higher the value of  $\mathit{W_{Penguin}(x)}$.  Hence, provided the relevant properties of penguins (such as non-flying) remain unaltered, the more typical is $x$ as a bird,  
the more typical is $x$ as a Penguin.
Notice also that the weight $\mathit{W_{Bird}(x)}$ of a domain element $x$ with respect to $\mathit{Bird}$ is related to the interpretation of $Bird$ in $I$ by the faithfulness condition or by a coherence condition (depending on the semantic construction).

\normalcolor

\section{A multi-preferential fuzzy interpretation of multilayer perceptrons} \label{sec:multiulayer_perceptron}

In this section, we first shortly introduce multilayer perceptrons. Then we develop a fuzzy multi-preferential interpretation of a neural network,
which can be used for post-hoc explanation, based on a model checking approach.

\subsection{Multilayer perceptrons}  \label{sec:MLPs}

Let us first recall from \cite{Haykin99} the model of a {\em neuron} as an information-processing unit in an artificial neural network. The basic elements are the following:
\begin{itemize}
\item 
a set of {\em synapses} or {\em connecting links}, each one characterized by a {\em weight}; we let $x_j$ be the signal at the input of synapse $j$ connected to neuron $k$, and $w_{kj}$ the related synaptic weight;
\item
the adder for summing the input signals to the neuron, weighted by the respective synapses weights: $\sum^n_{j=1} w_{kj} x_j$;
\item
an {\em activation function} for limiting the amplitude of the output of the neuron (typically, to the interval $[0,1]$ or $[-1,+1]$).
\end{itemize}
The logistic, threshold and hyperbolic-tangent functions are examples of activation functions.
A neuron $k$ can be described by the following pair of equations: $u_k= \sum^n_{j=1} w_{kj} x_j $, and $y_k=\varphi(u_k + b_k)$,
where  $x_1, \ldots, x_n$ are the input signals and $w_{k1}, \ldots,$ $ w_{kn} $ are the weights of neuron $k$;  
$b_k$ is the bias, $\varphi$ the activation function, and $y_k$ is the output signal of neuron $k$.
By adding a new synapse with input $x_0=+1$ and synaptic weight $w_{k0}=b_k$, one can write: \begin{equation} \label{eq:neuron2}
u_k= \sum^n_{j=0} w_{kj} x_j \mbox{ \ \ \ \ \ \ \ \ \ \ \ \ \ \ } y_k=\varphi(u_k),
\end{equation}
where $u_k$ is called the {\em induced local field} of the neuron.

A neural network can then be seen as ``a directed graph consisting of nodes with interconnecting synaptic and activation links"  \cite{Haykin99}:
nodes in the graph are the neurons (the processing units) 
and the weight $w_{ij}$ on the edge from node $j$ to node $i$ represents ``the strength of the connection [..]\ by which unit $j$ transmits information to unit $i$" \cite{Plunkett98}.
Source nodes (i.e., nodes without incoming edges) produce the input signals to the graph. 
Neural network models are classified by their synaptic connection topology. In a {\em feedforward} network the {architectural graph} is acyclic, while in a {\em recurrent} network it contains cycles. In a feedforward network neurons are organized in layers. In a {\em single-layer} network there is an input layer of source nodes and an output layer of computation nodes. In a {\em multilayer feedforward} network there are one or more hidden layers, whose computation nodes are called {\em hidden neurons} (or hidden units).
The source nodes in the input layer supply the activation pattern ({\em input vector}) providing the input signals for the first layer computation units.
In turn, the output signals of first layer computation units provide the input signals for the second layer computation units, and so on, up to the final output layer of the network, which provides the overall response of the network to the activation pattern.
In a recurrent network at least one feedback 
exists, so that ``the output of a node in the system influences in part the input applied to that particular element"  \cite{Haykin99}.   
In the following, we do not put restrictions on the topology the network, even though in Section \ref{sec:experimentation} we only report experiments on feedforward networks.

``A major task for a neural network is to learn a model of the world"  \cite{Haykin99}. In supervised learning, a set of input/output pairs, input signals and corresponding desired response, referred as training data, or training sample, is used to train the network to learn. In particular, the network learns by changing the synaptic weights, through the exposition to the training samples. After the training phase, in the generalization phase, the network is tested with data not seen before. ``Thus the neural network not only provides the implicit model of the environment in which it is embedded, but also performs the information-processing function of interest"  \cite{Haykin99}.
In the next section, we aim to make this model explicit as a multi-preferential model.

\subsection{A multi-preferential interpretation of MLPs and property verification by model checking}  \label{sec:fuzzy_sem_for_NN}

In this section, we show that a fuzzy multi-preferential interpretation (an $\alcFt$ interpretation) can be associated to a multilayer network $\enne$, based on the activity of the network over a set of input stimuli $\Delta$. 
Fuzzy and typicality properties of the network can then be verified by model checking over such an interpretation, and used for post-hoc explanation.

Assume that the network ${\cal N}$ has been trained and the synaptic weights $w_{kj}$ have been learned.
We associate a concept name $C_i \in N_C$ to the units  
$i$ of interest in ${\cal N}$, which may include input, output or hidden units. They are the units we are interested in, for property verification. 

We construct a multi-preferential interpretation over a (finite) {\em domain $\Delta$} of input stimuli;
for instance, the input vectors considered so far, for training and generalization, or a subset of it (e.g., the test set). 
In case the network is not feedforward, we assume that, for each input vector $v$ in $\Delta$, the network reaches a stationary state  \cite{Haykin99}, in which $y_k(v)$ is the activity level of unit $k$, 
and equations (\ref{eq:neuron2}) hold,  for all units $k$. 
We also assume the activation of units to be in the interval $[0,1]$.

Let $\Delta$ be a finite (non-empty) set of input vectors.
We can associate to ${\cal N}$ a fuzzy multi-preferential interpretation over $\Delta$, in the boolean fragment of $\alcFt$, 
which contains no roles (i.e., $N_R= \emptyset$) and no individual names (i.e., $N_I= \emptyset$).
We refer to the definition of an $\alcFt$ interpretation (Definition \ref{def:alcFt_interpretation}).

\begin{definition} \label{sec:model_of_a_network}
The  {\em fuzzy multi-preferential  interpretation 
of a network ${\cal N}$} over a non-empty domain $\Delta$,
is the $\alcFt$ interpretation $I_{\enne}^\Delta=\langle \Delta, \cdot^I \rangle$ where:
the interpretation function $\cdot^I$ satisfies the condition that, 
for all concept names $C_k \in N_C$ and for all $ x \in \Delta$,
$$C_k^I(x)= y_k(x)$$ 
where $y_k(x)$ is the output signal of neuron $k$, for input vector $x$.
\end{definition}
As we have seen in section  \ref{sec:fuzzyalc+T}, the $\alcFt$ interpretation $I_{\enne}^\Delta$ is a multi-preferential interpretation, as the fuzzy interpretation of concepts induces a preference relation associated to each concept. Here, the preferences associated with concepts are those associated with units, and based on the unit activations for the different inputs. More precisely, the preference relation
 $<_{C_k}$ associated to concept $C_k$ (and to unit $k$),  induced by the interpretation $I_{\enne}^\Delta$, is determined by the activity of unit $k$ as follows: for $x,x' \in \Delta$,
\begin{equation} \label{def:pref_M_N}
 x <_{C_k} x' \mbox{ iff } y_k(x) > y_k(x').
\end{equation}
This allows the set of typical instances of a concept $C_k$ to be identified according to the definition of typicality concepts in Equation (\ref{eq:interpr_typicality}), by selecting the input stimuli $x \in \Delta$ with the highest 
activity value $y_k(x)$.

This model provides a multi-preferential interpretation of the network $\enne$, 
based on the input stimuli considered in $\Delta$.
For instance, in case the neural network is used for categorization and an output neuron is associated to each category, each concept $C_h$ associated to an output unit $h$  corresponds to a learned category. 
If $C_h \in N_C$, the preference relation $<_{C_h}$ determines the relative typicality of input stimuli with respect to category $C_i$. This allows to verify typicality properties concerning categories,  such as $\tip(C_h) \sqsubseteq D \geq \alpha$ (where $D$ is a boolean concept built from the named concepts in $N_C$), by {\em model checking} on the model $I_{\enne}^\Delta$. According to the semantics of typicality concepts, this would require to identify typical $C_h$-elements 
and checking whether they are instances of concept $D$ with a degree greater than $\alpha$. 

For instance, in Section  \ref{sec:experimentation} we consider some example neural networks, trained to recognize emotions ({\em surprise, fear, happiness, anger}) in images of human faces.
In that case, we will be interested in understanding which properties have been learned by the network,
concerning the relationships between 
some learned category (e.g., happiness)
and some specific features of the image (in the example, facial muscle contractions). To this purpose, we will check properties such as,  for instance, 
$\tip(happiness) \sqsubseteq \mathit{au12}  \geq \alpha$ (where $au12$ is the activation of the lip corner puller muscle used for smiling), to verify whether
the images recognized by the network as typical instances of happy faces correspond to smiling faces, to some degree.

In general, fuzzy typicality inclusions of the form $\tip(C) \sqsubseteq D \theta \alpha$, with $C$ and $D$ boolean concepts, can be verified on the model $I_{\enne}^\Delta$ in polynomial time in the size of the model $I_{\enne}^\Delta$ and in the size of the formula.

Consider, for instance,  the verification of $\tip(C) \sqsubseteq D \geq \alpha$ under the choice of combination functions as in G\"odel logic.
The verification amounts  to check that
$\inf_{x \in \Delta}  \tip(C)^{I_{\enne}^\Delta}(x) \rhd D^{I_{\enne}^\Delta}(x) \geq \alpha$, i.e., that 
for all $x \in \Delta$, $ \tip(C)^{I_{\enne}^\Delta}(x) \rhd D^{I_{\enne}^\Delta}(x) \geq \alpha$ holds.
When  $\tip(C)^{I_{\enne}^\Delta}(x) =0$, that is, $x$ is not a typical $C$-element, 
$ \tip(C)^{I_{\enne}^\Delta}(x) \rhd D^{I_{\enne}^\Delta}(x) \geq \alpha$ holds trivially.

The identification of typical $C$-elements in $\Delta$ requires: 
computing the values of $C^{I_{\enne}^\Delta}(x)$, for all input stimuli $x \in \Delta$ and selecting those $y$ such that 
the value $C^{I_{\enne}^\Delta}(y)$ is maximal among the values of $C^{I_{\enne}^\Delta}(x)$, for all $x \in \Delta$.
Then, for all typical $C$-elements $x$,  
one has to verify that $ C^{I_{\enne}^\Delta}(x) \rhd D^{I_{\enne}^\Delta}(x) \geq \alpha$ holds, which
requires to verify that $C^{I_{\enne}^\Delta}(x)  \leq D^{I_{\enne}^\Delta}(x)$ or   $D^{I_{\enne}^\Delta}(x) \geq \alpha$ hold.
In turn,  this requires the value of $D^{I_{\enne}^\Delta}(x)$  to be computed, for all typical $C$-elements $x$.

Overall, the verification requires a polynomial number of steps in the size of the model $I_{\enne}^\Delta$ and in the size of the formula $\tip(C) \sqsubseteq D$. 
Note that, as $C$ and $D$ only contain a polynomial number of subformulas,  the values of $C(x)$ and $D(x)$, for some $x \in \Delta$ can be computed in polynomial time. But the evaluation has to be repeated for all elements $x \in \Delta$, and the domain $\Delta$ can be very large.

It is easy to see that similar polynomial algorithms can be developed for the verification of inclusions of the form  $\tip(C) \sqsubseteq D \leq \alpha$ (which require the verification that there is an element $x \in \Delta$,  such that $ \tip(C)^{I_{\enne}^\Delta}(x) \rhd D^{I_{\enne}^\Delta}(x) \leq \alpha$ holds), and for the verification of strict inclusions $C \sqsubseteq D \theta \alpha$,  according to the choice of the t-norm, s-norm, negation and implication functions.
In general, inclusion axioms of the form $C \sqsubseteq D \theta \alpha$ may be considered, where $C$ and $D$ contain (non-nested) occurrences of the typicality operator $\tip$.
\begin{proposition}
Whether an axiom $C \sqsubseteq D \theta \alpha$ is satisfied in a multi-preferential interpretation $I_{\enne}^\Delta$, can be decided in polynomial time in the size of $I_{\enne}^\Delta$ and in the size of $C \sqsubseteq D$.
\end{proposition}

The size of model $I_{\enne}^\Delta$ is $O(|N_C| \times |\Delta|)$: it depends on the 
number  $|N_C|$ of the units in the network that we are considering for property verification,
and on the size of the set of input stimuli $\Delta$, which can be very large. 
Observe, however, that to prove an inclusion $ \tip(C) \sqsubseteq D \theta  \alpha $ (or $ C \sqsubseteq D \theta  \alpha $) we do not need to consider and build  the entire model $I_{\enne}^\Delta$, but it is sufficient to consider the restriction of the model over the concept names in $C$ and in $D$, as only the interpretation of the subconcepts occurring in $C$ and in $D$ 
are needed in the verification.

In Section  \ref{sec:experimentation} we report results of the model checking approach in the verification of typicality properties of a multilayer networks, trained to recognize emotions from input features, exploiting a Datalog encoding of the model checking problem developed in  \cite{Datalog2022} for the finite-valued case.

\subsection{Multilayer perceptrons as weighted conditional knowledge bases} \label{sec:NN&Conditionals}  

Another possible approach for reasoning about the properties of a neural network consists in exploiting entailment in the defeasible logic, based on the idea that the neural network  ${\enne}$ can be regarded as a defeasible knowledge base $K_{\enne}$. In this section, we explore this approach.

Let us introduce a concept name $C_i \in N_C$ for each unit $i$ in the network $\enne$ 
and let ${\cal C}= \{ C_1, \ldots, C_n\}$ be a subset of $N_C$, namely the set of all concept names  $C_i \in N_C$ such that there is at least a synaptic connection
between some unit $j$ and unit $i$.
Given the {\em fuzzy multi-preferential interpretation} $I_{\enne}^\Delta=\langle \Delta, \cdot^I \rangle$  as defined in Section  \ref{sec:fuzzy_sem_for_NN},
we aim at proving that $I_{\enne}^\Delta$ is indeed a model of the neural network ${\enne}$ in a logical sense.

A weighted conditional knowledge base $K^{\enne}$ can be defined from the neural network ${\enne}$ as follows.
For each unit $k$ with incoming edges, we consider all the units $j_1, \ldots, j_m$ whose output signals are the input signals   
of unit $k$, with synaptic weights $w_{k,{j_1}}, \ldots, w_{k,{j_m}}$.  
Let $C_k$ be the concept name associated to unit $k$ and 
$C_{j_1}, \ldots, C_{j_m}$ be the concept names associated to units $j_1, \ldots, j_m$, respectively.
We define for each concept $C_k \in {\cal C}$ a set ${\cal T}_{C_k}$ of typicality inclusions, with their associated weights, as follows:
\begin{quote}
$\tip(C_k) \sqsubseteq C_{j_1}$ with  $w_{k,{j_1}}$, \\
$\ldots$ ,\\
$\tip(C_k) \sqsubseteq C_{j_m}$ with  $w_{k,{j_m}}$
\end{quote}
The knowledge base  constructed from network ${\enne}$ is defined, from the above set ${\cal C}$ of distinguished concepts, as the tuple: $K^{\enne} = \langle  {\cal T}_{f},{\cal T}_{C_1}, \ldots,$ $ {\cal T}_{C_n}, {\cal A}_f  \rangle$, where $ {\cal T}_{f}=\emptyset$, ${\cal A}_f=\emptyset$ and,  for each $C_k \in {\cal C}$, ${\cal T}_{C_k}$ is the set of weighted typicality inclusions associated to neuron $k$ as defined above.

$K^{\enne}$ is a weighted knowledge base over the set  of distinguished concepts ${\cal C}= \{ C_1, \ldots, C_n\}$.
For multilayer feedforward networks, $K^{\enne}$ corresponds to an acyclic conditional knowledge base, 
and defines a (defeasible) subsumption hierarchy among concepts. 
In the more general case, when the network may contain cycles, our characterization is intended to capture the properties of stationary states of the network \cite{Haykin99}.
We prove that, when a concept name $C_k$ is introduced for each unit $k$ in the network $\enne$, the multi-preferential interpretation $I_{\enne}^\Delta$, defined in Section  \ref{sec:fuzzy_sem_for_NN}, is a $\varphi$-coherent multi-preferential model of the weighted knowledge base $K^{\enne}$.
 
Let us refer to the network $\enne$ above, in which the output signals of units $j_1, \ldots, j_m$  are the input signals   
of unit $k$ with synaptic weights $w_{k,{j_1}}, \ldots, w_{k,{j_m}}$, respectively.
The intuition is that, as concept name $C_k$ is associated to unit $k$ in  $\enne$ and concept names $C_{j_h}$ are associated to each unit $j_h$, the following holds: $C_k^I(x)$ corresponds to the activation $y_k$ of unit $k$ for a given input stimulus $x$, while $C_{j_h}^I(x)$ corresponds to the activation $y_{j_h}$ of unit $j_h$ for the same stimulus.
Hence, the sum  $ \sum^m_{h=0} w_{k,{j_h}} C_{j_h}^I(x)$ corresponds to the {\em induced local field} $u_k$ of neuron $k$,
and equation $y_k=\varphi(u_k)$ in (\ref{eq:neuron2}) (which holds for a stationary state, in non-feedforward networks), enforces the $\varphi$-coherence condition
 $C_k^I(x) =\varphi( W_k(x))$, where $\varphi$ is the activation function of unit $k$.

Let $I_{\enne}^\Delta=\langle \Delta, \cdot^I \rangle$  be the fuzzy multi-preferential interpretation of network $\enne$ over a domain $\Delta$ of input stimuli, as defined in Section  \ref{sec:fuzzy_sem_for_NN}. Assume  that $N_C$ contains a concept name $C_i$ for each unit $i$ in the network. 
We can prove the following proposition.

\begin{proposition} \label{prop:I^enne_varphi_coherent_model}
$I_{\enne}^\Delta$ is a $\varphi$-coherent multi-preferential model of the weighted knowledge base $K^{\enne}$.
\end{proposition}
\begin{proof}
Let $\enne$ be network such that $\varphi_i$ is the activation function of unit $i$ in $\enne$.  
Let $K^{\enne}$ be the weighted knowledge base over the set  of distinguished concepts ${\cal C}= \{ C_1, \ldots, C_n\}$, associated to $\enne$ as in the construction above.

Let the fuzzy multi-preferential interpretation $I_{\enne}^\Delta=\langle \Delta, \cdot^I \rangle$  of $\enne$ over a domain $\Delta$  be defined according to Definition  \ref{sec:model_of_a_network}, in Section  \ref{sec:fuzzy_sem_for_NN},
but assuming that $N_C$ contains a concept name $C_i$ for each unit $i$ in the network. 

Given the set ${\cal T}_{C_k}$ of weighted typicality inclusions for $C_k \in {\cal C}$ in $K^{\enne}$:
\begin{quote}
$\tip(C_k) \sqsubseteq C_{j_1}$ with  $w_{k,{j_1}}$, \\
$\ldots$ ,\\
$\tip(C_k) \sqsubseteq C_{j_m}$ with  $w_{k,{j_m}}$
\end{quote}
by construction, there are units $k, j_1, \ldots, j_m$ in $\enne$, such that 
the output signals of units $j_1, \ldots, j_m$  are the input signals   
of unit $k$ with synaptic weights $w_{k,{j_1}}, \ldots, w_{k,{j_m}}$.

By construction of the fuzzy interpretation $I_{\enne}^\Delta$, for all $x \in \Delta$ and $C_k \in N_C$, $C_k^{I_{\enne}^\Delta}(x)=y_k(x)$,
i.e., $C_k^{I_{\enne}^\Delta}(x)$ corresponds to the activation $y_k(x)$ of neuron $k$ for the stimulus $x$.
We have to prove that $I_{\enne}^\Delta$ satisfies the $\varphi$-coherence condition.

Note that, in the construction of $I_{\enne}^\Delta$, in case the network is not feedforward, we have  assumed that, for any input stimulus $x$ in $\Delta$, the network reaches a stationary state, in which (for all $k$) $y_k(x)$ is the activity level of unit $k$. 
Then, equations (\ref{eq:neuron2}) holds  for unit $k$, i.e.:
\begin{align*}
u_k= \sum^m_{h=0} w_{k,{j_h}} y_{j_h} \mbox{ \ \ \ \ \ \ \ \ \ \ \ \ \ \ } y_k=\varphi_k(u_k),
\end{align*}
where $\varphi_k$ is the activation function of unit $k$.
Making input $x$ explicit,  it must hold that:
$y_k(x)= \varphi_k(\sum^m_{h=0} w_{k,{j_h}} y_{j_h}(x))$,
 that is to say:
 \begin{align*}
 C_k^I(x) =\varphi_k( \sum^m_{h=0} w_{k,{j_h}} C_{j_h}^I(x))
 \end{align*}
 As the equation above holds for all concepts $C_k \in {\cal C}$, and each domain element $x\in \Delta$, the interpretation $I_{\enne}^\Delta$ 
 satisfies the $\varphi$-coherence condition and is  
 a $\varphi$-coherent model of   $K^{\enne}$.
\end{proof}

The next corollaries follow from Proposition \ref{prop:I^enne_varphi_coherent_model} and Proposition \ref{prop:phi_coherent_models}, under the assumptions of Proposition \ref{prop:I^enne_varphi_coherent_model}, that is: $I_{\enne}^\Delta$  is a fuzzy multi-preferential interpretation of a network $\enne$ built over a domain $\Delta$ of input stimuli, as defined in Section  \ref{sec:fuzzy_sem_for_NN}, and  $N_C$ contains a concept name $C_i$ for each unit $i$ in $\enne$.

\begin{corollary}   \label{Prop_faithful}
$I_{\enne}^\Delta$ is a faithful multi-preferential model of the weighted knowledge base $K^{\enne}$, provided the activation functions $\varphi_k$ of all units are monotone non-decreasing.
\end{corollary}

\begin{corollary} \label{Prop:coherent_model}
$I_{\enne}^\Delta$ is a coherent multi-preferential model of the weighted knowledge base $K^{\enne}$, provided the activation functions $\varphi_k$ of all units are monotonically increasing. 
\end{corollary}
Corollary  \ref{Prop:coherent_model} simplifies the formulation of 
Proposition 1 in \cite{JELIA2021}. Unlike in \cite{JELIA2021}, here we are considering a non-crisp interpretation for typicality concepts.

By Proposition 
\ref{prop:I^enne_varphi_coherent_model}
the interpretation  $I_{\enne}^\Delta$ constructed from the network $\enne$,
by considering the activations of units over the input stimuli in $\Delta$,  is a model of the network in a logical sense, as it is a $\varphi$-coherent model of the conditional  knowledge base $K^{\enne}$ associated to the network.

We can prove that, under the $\varphi$-coherent semantics,  the knowledge base $K^{\enne}$ provides a logical characterization of the neural network $\enne$,  
as the following also holds:
given any $\varphi$-coherent model $I=\langle \Delta, \cdot^I \rangle$ of the knowledge base $K^{\enne}$, each domain element $x \in \Delta$ corresponds to a stationary state of the network $\enne$, that is, equations (\ref{eq:neuron2}) are satisfied when the activity level $y_k$ of unit $k$ is taken to be the value $C_k^I(x)$, for each $k$.

\begin{proposition}\label{prop:varphi_coherent_model_to_enne}
Let  $K^{\enne}$ be the weighted knowledge base associated to a multilayer network $\enne$.
Let $I=\langle \Delta, \cdot^I \rangle$ be any $\varphi$-coherent model of $K^{\enne}$.
For all $x \in \Delta$, let  $y_j=C_j^I(x)$ be the output signal of unit $j$, for each unit $j$. Then, equations (\ref{eq:neuron2}) hold for any unit $k$ with incoming edges.
\end{proposition}
\begin{proof}
Consider an element $x \in \Delta$, and let  $k$ be a unit with incoming edges such that the output signals of units $j_1, \ldots, j_m$  are the input signals   
of unit $k$ with synaptic weights $w_{k,{j_1}}, \ldots, w_{k,{j_m}}$.

By construction of $K^{\enne}$, from the $\varphi$-coherence condition, it must hold that:
\begin{align*}
C_k^I(x) =\varphi_k( \sum^m_{h=0} w_{k,{j_h}} C^I_{j_h}(x)),
\end{align*}
Hence, $C_k^I(x) =\varphi_k( u_k)$, and  $u_k=\sum^m_{h=0} w_{k,{j_h}} C^I_{j_h}(x)$.

As from the hypothesis $y_k=C_k^I(x)$ and,  for all $h$, $y_{j_h}=C_{j_h}^I(x)$ it holds:
\begin{align*}
u_k= \sum^m_{h=0} w_{k,{j_h}} y_{j_h} \mbox{ \ \ \ \ \ \ \ \ \ \ \ \ \ \ } y_k=\varphi_k(u_k),
\end{align*}
That is, equations  (\ref{eq:neuron2}) are satisfied.
\end{proof}

Let us observe that any {\em canonical}  $\varphi$-coherent model of $K^{\enne}$ contains all the stationary states of the network $\enne$. 
For feedforward networks, a canonical model describes the activity of all units in the network for all the (possibly infinitely many) input stimuli. 

Proof methods for reasoning in the $\varphi$-coherent multi-preferential semantics have been developed in \cite{ICLP22,arXiv2023}, for the fragment $\lc$ of $\alc$ without roles and role restrictions, based on the finitely many-valued G\"odel description logic or  \L ukasiewicz description logic, extended with typicality. More precisely,
an Answer Set Programming encoding of  an approximation of $\varphi$-coherent entailment  (called $\varphi_n$-coherent entailment)
has been developed for the boolean fragment $\lcnt$ of $\lc$ plus typicality, over the truth space  $\{0, \frac{1}{n},\ldots, \frac{n-1}{n}, \frac{n}{n}\}$, for an integer $n \geq 1$.
The study of the finitely-valued case, is indeed motivated by the undecidability results for fuzzy description logics with general inclusion axioms \cite{CeramiStraccia2013,BorgwardtPenaloza12}. 

In the next section, we prove that, under suitable conditions, the $\varphi$-coherent semantics  in the finitely-valued case is indeed an approximation of the 
$\varphi$-coherent semantics in the fuzzy case.

\section{Approximating $\varphi$-coherent models in the finitely-valued case} \label{sec:finitely_valued_case}

While in Sections  \ref{sec:fuzzyalc+T} and  \ref{sec:closure} we have defined a fuzzy $\alc$ with  typicality and its closure constructions, in a similar way, one can define a {\em finitely many-valued  $\alc$ with typicality}, ${\alc}_n \tip$, by building on the finitely-valued description logic ${\alc}_{n}$, and taking ${\cal C}_n= \{0, \frac{1}{n},\ldots,$ $ \frac{n-1}{n}, \frac{n}{n}\}$ (for $n \geq 1$) as the truth value space. 

The idea is that of approximating function $\varphi$ with a function $\varphi_n$ over the truth space ${\cal C}_n$, by developing a $\varphi_n$-coherent semantics,
which is indeed an approximation of the $\varphi$-coherent semantics (under some conditions). 
In the following, for simplicity, we consider a single function $\varphi$, rather than a different function $\varphi_i$ for each unit $i$, but the results generalize to the case of multiple functions.

Let us assume that $\varphi$ is continuous function $\varphi: {\mathbb{R}} \rightarrow [0,1]$, and that the chosen t-norm, s-norm and negation function in $\lcFt$ are continuous as well. We define the $\varphi_n$-coherent semantics as follows.

Values $v \in [0,1] $ are approximated to the nearest value in ${\cal C}_n$:
\begin{align}\label{approx}
	[v]^n & = \left\{\begin{array}{ll}
						 0 & \mbox{ \ \ \ \  if } v \leq \frac{1}{2n} \\
						 \frac{i}{n} & \mbox{ \ \ \ \  if } \frac{2i -1}{2n} <  v \leq \frac{2i +1}{2n}, \mbox{ for } 0<i<n \\
						1 &  \mbox{ \ \ \ \  if } \frac{2n -1}{2n} <  v
					\end{array}\right.
\end{align} 

For an integer $n\geq 1$, let  $\varphi_n: {\mathbb{R}} \rightarrow {\cal C}_n$ 
be defined as: 
$$\varphi_n(z) = [\varphi(z)]^n, $$
for all $z \in {\mathbb{R}}$.
The notions of   {\em $\varphi_n$-coherent model} and {\em $\varphi_n$-coherent entailment} can be defined  
similarly to {\em $\varphi$-coherent model} and {\em $\varphi$-coherent entailment},
by replacing $\varphi$ with $\varphi_n$ in Definitions 
\ref{def:varphi-model}  and \ref{fm-entailment}.

Observe that the sequence of functions $(\varphi_n)_{n \in {\mathbb{N}}}$ uniformly converges to function $\varphi$, i.e., 
for all $\varepsilon>0$ there is an $n_0 \in  {\mathbb{N}}$ such that, for all $n \geq n_0$, 
\begin{align}\label{eqA}
\mid  \varphi_n(z) -  \varphi(z) \mid < \varepsilon, \mbox{ $\forall z \in {\mathbb{R}}$ }
\end{align}
Indeed, from the definition of $\varphi_n$, $| \varphi_n(z) - \varphi(z)|\leq \frac{1}{2n}$. 
We can get $| \varphi_n(z) - \varphi(z)|\leq  \frac{1}{2n} < \varepsilon$, by choosing $n_0=\lceil \frac{1}{2\varepsilon} \rceil +1$.

For any $v \in  {\mathbb{R}}$, $lim_{n \ri \infty}[v]^n=  v$.
Hence, for any concept name $A\in N_C$, fuzzy $\lcFt$ interpretation $I=\la \Delta, \cdot^I\ra$ and $x \in \Delta$,
$lim_{n \ri \infty}[A^I(x)]^n=  A^I(x)$.
As we are considering continuous combination functions, for any concept $D_j$,
$lim_{n \ri \infty}[D_j^I(x)]^n=  D_j^I(x)$.
Let $W_i^I(x)= \sum_{h} w_h^i  \; D_{i,h}^I(x)$ and let $W_i^{I,n}(x)= \sum_{h} w_h^i  \; [D_{i,h}^I(x)]^n$.
As $W_i^{I,n}(x)$ is continuous in $[D_{i,1}^I(x)]^n,$ $ \ldots, [D_{i,k}^I(x)]^n$, and  $\varphi$ is as well continuous, their composition is a continuous function, and:
\begin{align} \label{eqB}
lim_{n \ri \infty} \varphi(W_i^{I,n}(x)) =  \varphi(W_i^I(x))
\end{align}
that is, for all $\varepsilon>0$ there is an $m_0 \in  {\mathbb{N}}$ such that, for all $n \geq m_0$, 
$$\mid\varphi(W_i^{I,n}(x))  -   \varphi(W_i^I(x))  \mid < \varepsilon.$$
Therefore the following lemma holds.
\begin{lemma} \label{Lemma1}
Given a continuous 
function $\varphi: {\mathbb{R}} \rightarrow [0,1]$, and an $\lcFt$ interpretation $I$,
$lim_{n \ri \infty} \varphi_n(W_i^{I,n}(x))=  \varphi(W_i^I(x))$, for all $i=1,\ldots,k$.
\end{lemma}

%
Given an $\lcFt$ interpretation $I=\la \Delta, \cdot^I\ra$, we can define an $\lcFtn$ interpretation $I_n=\la \Delta, \cdot^{I_n}\ra$ over the value space ${\cal C}_n$ by letting:
$C^{I_n}(x)=[C^I(x)]^n$, for all concepts $C$, and  $a^{I_n}= a^I$, for all $a \in N_I$.

\noindent
We can then prove the following proposition.
\begin{proposition}
 Let $K=\langle  {\cal T},$ $ {\cal T}_{C_1}, \ldots,$ $ {\cal T}_{C_k}, {\cal A}  \rangle$ be  a weighted $\lcFt$  knowledge base,
and 
$\varphi: {\mathbb{R}} \rightarrow [0,1]$ a continuous function.
\begin{itemize}
\item[(i)]
If 
$C_i^I(x)= \varphi  (\sum_{h} w_h^i  \; D_{i,h}^I(x)) $,
then, for all $\varepsilon >0$, there is a  $k_0 \in  {\mathbb{N}}$ such that for all $n\geq k_0$, 
$|C_i^{I_n}(x) - \varphi_n (\sum_{h} w_h^i  \; D_{i,h}^{I_n}(x))| <  \varepsilon$.

\item[(ii)]
If 
$C_i^I(x) \neq \varphi  (\sum_{h} w_h^i  \; D_{i,h}^I(x)) $,
then there exist an $\varepsilon >0$ and a  $k_0 \in  {\mathbb{N}}$ such that for all $n\geq k_0$, 
$|C_i^{I_n}(x) - \varphi_n (\sum_{h} w_h^i  \; D_{i,h}^{I_n}(x))|> \varepsilon$.

\end{itemize}
\end{proposition}
 \begin{proof}
 For item (i),
assume 
condition
$C_i^I(x)= \varphi  (\sum_{h} w_h^i  \; D_{i,h}^I(x)) $
holds. 
As $C_i^{I_n}(x)$ converges to $C_i^I(x)$, and, by Lemma \ref{Lemma1}, $\varphi_n (\sum_{h} w_h^i  \; D_{i,h}^{I_n}(x))$ converges to $\varphi  (\sum_{h} w_h^i  \; D_{i,h}^I(x)) $, the thesis follows.

For item (ii), assume $C_i^I(x) \neq \varphi  (\sum_{h} w_h^i  \; D_{i,h}^I(x)) $, and let 
$d=|C_i^I(x) - \varphi  (\sum_{h} w_h^i  \; D_{i,h}^I(x)) |$.
Let $\varepsilon = d/3$.

As $C_i^{I_n}(x)$ converges to $C_i^I(x)$, there is an $n_0 \in  {\mathbb{N}}$ such that for all $n\geq n_0$, 
$|C_i^{I_n}(x) - C_i^I(x)|< \varepsilon =d/3$.
 By Lemma \ref{Lemma1}, there is an $m_0 \in  {\mathbb{N}}$ such that for all $n\geq m_0$,  
 $|\varphi_n (\sum_{h} w_h^i  \; D_{i,h}^{I_n}(x))- \varphi  (\sum_{h} w_h^i  \; D_{i,h}^I(x)) |< \varepsilon =d/3$.
  
Let $k_0= max\{n_0,m_0\}$.
Then, 
\begin{center}
$|C_i^{I_n}(x) - \varphi_n (\sum_{h} w_h^i  \; D_{i,h}^{I_n}(x)| \geq d/3=\varepsilon$
\end{center}
for all $n \geq k_0$.
\end{proof}

Note that the notion of $\varphi_n$-coherence may fail to characterize all the  stationary states of a network as, although for some $x \in \Delta$ it may hold that $C_i^I(x)= \varphi  (\sum_{h} w_h^i  \; D_{i,h}^I(x)) $, for all concepts $C_i$,
there is no guarantee that some $n_0$ exists such that, for all $n >n_0$,
 $C_i^{I_n}(x) = \varphi_n (\sum_{h} w_h^i  \; D_{i,h}^{I_n}(x))$.  
 Nevertheless, by item $(i)$, at the limit the distance between  $C_i^{I,n}(x)$ and $ \varphi_n (\sum_{h} w_h^i  \; D_{i,h}^{I,n}(x))$ converges to $0$.

On the other hand, it may be the case that $C_i^{I_n}(x) = \varphi_n (\sum_{h} w_h^i  \; D_{i,h}^{I_n}(x))$ holds for some $n$, due to the approximation,
while $C_i^I(x) = \varphi  (\sum_{h} w_h^i  \; D_{i,h}^I(x)) $ does not hold. In such a case, by item $(ii)$, there must be a $k_0$ such that for all values of $n \geq k_0$, the first equality will not hold.

 In the following section we exploit the proof methods developed in \cite{arXiv2023,Datalog2022}
in the verification of the properties of some trained feedforward networks under the $\varphi_n$-coherent semantics.


\section{An experimentation: model checking and entailment for the verification of facial emotion recognition} \label{sec:experimentation}

While a neural network, once trained, can quickly classify new stimuli (i.e., perform instance checking), other reasoning services such as satisfiability, entailment and model-checking are missing. Such reasoning tasks are useful for validating knowledge that has been learned, including proving whether the network satisfies some (strict or conditional) properties. 

In the  finitely-valued case,  
Datalog with weakly stratified negation has been used for developing a model-checking approach for verifying multilayer networks  \cite{Datalog2022}.
Still in the finitely-valued case,
an ASP-based approach can be exploited for reasoning with weighted conditional KBs under $\varphi_n$-coherent entailment \cite{ICLP22,arXiv2023}.

Both the entailment and the model-checking approaches have been experimented in the verification of properties of some trained feedforward networks and, in the following, we report some results. 

We concentrate, in particular, 
on the verification of formulae
of the form $\tip(E) \sqsubseteq F \geq \alpha$ where $E$ is an output class
(i.e.\ one of the possible outputs of
classification, or a single output class
the network is trained to recognize),
and $F$ is a boolean combination
of input classes.

The interest for such formulae lies in the fact that a property
$\tip(E) \sqsubseteq F \geq \alpha$
tells something about the stimuli that
are classified as $Es$ with high membership
(highest in ${\cal C}_n$), and could then be seen
as describing what the network intends
as a prototypical $E$. 

It might of course be
the case that $\tip(E) \sqsubseteq F \geq \alpha$ holds, and the corresponding strict version
$E \sqsubseteq F \geq \alpha$
does not.
Similar considerations apply to inclusions of the form
$F \sqsubseteq \tip(E)$ and $F \sqsubseteq E$.

\subsection{Model checking} \label{sec:model_checking}

Based on the general idea of using model checking for verifying the properties of a neural network, as described in Section \ref{sec:multiulayer_perceptron}, in \cite{Datalog2022} we have developed a Datalog-based  approach which  builds a multi-valued preferential interpretation of a trained feedforward network ${\cal N}$ and, then, verifies the properties of the network for post-hoc explanation.

The Datalog encoding uses weakly stratified negation and contains a component $\Pi({\enne},\Delta,n)$ which is intended to build a (single) many-valued, preferential interpretation $I_{\enne}^\Delta$ with truth degrees in ${\cal C}_n$, and a component associated to the formulae to be checked. 

The model checking approach has been experimented  in the verification of properties of neural networks for the recognition of basic emotions  using the Facial Action Coding System (FACS) \cite{FACS}.
The RAF-DB \cite{rafdb} data set contains almost 30000 images labeled with basic emotions or combinations of two emotions. 
It was used as input to OpenFace 2.0 \cite{openface}, which detects a subset of the Action Units (AUs) in \cite{FACS}, i.e., facial muscle contractions. The relations between such AUs and emotions, studied by psychologists \cite{waller}, can be used as a reference for formulae to be verified on neural networks trained to learn such relations.

From the original dataset, the images labelled with a single emotion in the set $\{ surprise, fear, happiness, anger \}$ were selected. The dataset, with 4\,283 images, was highly unbalanced, then
the data was preprocessed by subsampling the larger classes and augmenting the minority ones using standard data augmentation techniques.
The processed dataset contains 5\,975 images.
The images were input to OpenFace 2.0;
the output intensities were rescaled in order to make their distribution conformant to the expected one in case AUs are recognized by humans \cite{FACS}.
The resulting 17 AUs were used as input to a neural network trained to classify its input as an instance of the four emotions. The neural network model used is a fully connected feed forward neural network with three hidden layers having 1\,800, 1\,200, and 600 nodes (all hidden layers use ReLU activation functions, while the softmax function is used in the output layer);
the F1 score of the trained network is 0.744
(the data quality was not very high; other emotions were not considered, in order to achieve
a reasonable accuracy).

The model checking approach in \cite{Datalog2022}
was easily adapted to the non-crisp notion of typicality we consider in this paper, and applied, using the Clingo ASP solver as Datalog engine, taking as set of input stimuli $\Delta$ the test set, containing 1\,194 images, and $n=5$, given that AU intensities, when assigned by humans, are on a scale of five values
(plus absence).
Table \ref{mcresults} reports some results for the verification of typicality inclusions $\tip(E) \sqsubseteq F \geq k/n$ in the finitely-valued G\"odel description logic with involutive negation plus typicality $G_n \lc \tip$, with the number of typical individuals for the emotion $E$,
and the number of counterexamples for different values of $k$\footnote{
In \cite{Datalog2022} the  conditional probability of concept $F$ given concept $\tip(E)$ has also been considered, based on Zadeh's probability of fuzzy events \cite{Zadeh1968}.}. 

For example, the inclusion axiom
$\tip(happiness) \sqsubseteq \mathit{au12}  \geq 3/5$ (where $au12$ is the activation of the lip corner puller muscle used for smiling) does not hold in the interpretation $I_{\enne}^\Delta$, since it has 1 counterexample out of 255 instances of $\tip(happiness)$, 
in fact, there is an instance $x$ such that $(\tip(happiness))^{I_{\enne}^\Delta}(x) \rhd au12^{I_{\enne}^\Delta}(x)  < 3/5$, given that  $(\tip(happiness))^{I_{\enne}^\Delta}(x) =1 $ and $ au12^{I_{\enne}^\Delta}(x)= 2/5$.
The property holds for $2/5$, i.e., $\tip(happiness) \sqsubseteq \mathit{au12}  \geq 2/5$ holds.
The formula
$\tip(happiness) \sqsubseteq  \mathit{au1} \sqcup \mathit{au6} \sqcup \mathit{au12} \sqcup \mathit{au14}  \geq 3/5$ also holds; the other action units involved are the activations of the inner brow raiser, cheek raiser, and dimpler.

The corresponding strict inclusions,
$happiness \sqsubseteq \mathit{au12}  \geq k/5$ and
\linebreak
$happiness \sqsubseteq  \mathit{au1} \sqcup \mathit{au6} \sqcup \mathit{au12} \sqcup \mathit{au14}  \geq k/5$,
do not hold even for $k=1$.

\begin{table}[t]
	\centering
	\begin{footnotesize}
	\begin{tabular}{ c || c || c  c  c  c  || c  }
		\hline
		\hline
		E \ \ &     F   &  k=1  & k=2 & k=3   & k=4 &  \#T(E)   \\	
		\hline
		\hline
	    $\mathit{surprise}$ 
		&  $ \mathit{au1} \sqcup \mathit{au2} \sqcup \mathit{au5} $ 
		& 54
		& 66
		& 79 
		& 140
		& 294
		\\
		\hline
	    $\mathit{surprise}$ 
		& $ \mathit{au1} \sqcup \mathit{au5} \sqcup \mathit{au15} \sqcup \mathit{au20} \sqcup \mathit{au26}  $ 
		& 2
		& 3
		& 6 
		& 59
		& 294
		\\
		\hline
		\hline
		$\mathit{fear}$
		& $ \mathit{au1} \sqcup \mathit{au2} \sqcup \mathit{au4} \sqcup \mathit{au5} $ 
		& 7
		& 9
		& 10 
		& 21
		& 45
		\\
		\hline
		$\mathit{fear}$
		& $ \mathit{au1} \sqcup \mathit{au2} \sqcup \mathit{au4} \sqcup \mathit{au5} \sqcup \mathit{au20} \sqcup \mathit{au26} $
		& 0
		& 0
		& 2 
		& 9
		& 45
		\\
		\hline
		\hline
		$\mathit{happiness}$
		&  $ \mathit{au1} \sqcup \mathit{au6} \sqcup \mathit{au12} \sqcup \mathit{au14} $
		& 0
		& 0
		& 0 
		& 22
		& 255
		\\
		\hline
		$\mathit{happiness}$
		& $ \mathit{au6} \sqcup \mathit{au12} $
		& 0
		& 0
		& 1
		& 32
		& 255
		\\	
		\hline
		$\mathit{happiness}$
		& $ \mathit{au6} \sqcap \mathit{au12} $
		& 6
		& 15
		& 23
		& 98
		& 255
		\\		
		\hline
		$\mathit{happiness}$
		& $ \mathit{au12} $
		& 0
		& 0
		& 1
		& 35
		& 255
		\\		
		\hline
		\hline
		$\mathit{anger}$
		&  $ \mathit{au4} \sqcup \mathit{au5} \sqcup \mathit{au7} \sqcup \mathit{au23} $
		& 5
		& 6
		& 7
		& 44
		& 212
		\\
		\hline
		\hline
	\end{tabular}
	\end{footnotesize}
	\caption{Results for checking formulae on the test set. \label{mcresults} The number of
	counterexamples for $\tip(E) \sqsubseteq F \geq k/n$ is provided for $k=1,\ldots,4$, as well as the total number of instances of $\tip(E)$.}
\end{table}

\subsection{Entailment} 
\label{sec:Entailment}

Based on the approximation of the $\varphi$-coherence semantics considered in section \ref{sec:finitely_valued_case} , 
Answer Set Programming (ASP) has been shown to be suitable for addressing defeasible reasoning in the finitely many-valued case with truth space ${\cal C}_n =\{0, \frac{1}{n},\ldots, \frac{n-1}{n}, \frac{n}{n}\}$ \cite{ICLP22}. 
A $\sc{P^{NP[log]}}$-completeness result for canonical $\varphi_n$-coherent entailment 
has been proven in \cite{arXiv2023} and some ASP encodings that deal with weighted knowledge bases with large search spaces have been developed.

The entailment of 
a typicality inclusion such as $\tip(C) \sqsubseteq D \theta \alpha$ from a weighted knowledge base $K$ is considered in the finitely-valued G\"odel description logic with involutive negation plus typicality
$G_n \lc \tip$, introduced in \cite{ICLP22} for the boolean fragment $\lc$ of $\alc$.
The verification can be formulated as a problem of computing {\em preferred answer sets} of an ASP program, considering a single distinguished individual, intended to represent a typical $C$-element, and selecting, as preferred answer sets, the ones maximizing the membership of the individual in concept $C$.
For the entailment problem, 
the upper bound in \cite{ICLP22} has been improved to ${\sc P}^{\sc NP[log]}$ by showing an algorithm running in polynomial time and performing \emph{parallel} queries to an NP oracle ($\sc P^{||NP}$)  \cite{arXiv2023}. The problem has also been shown to be ${\sc P}^{\sc NP[log]}$-complete
and the proof-of-concept ASP encoding has been redesigned so to obtain the desired multi-preferential semantics by taking advantage of weak constraints.
The scalability of the different ASP encodings has been assessed empirically.

The entailment approach has been experimented on the same domain as the model checking approach,
for a binary classification task, for the class 
$happiness$
vs other emotions. A set of 8\,835 images was used (no augmentation was needed in this case for balancing).
The images were input, as in the previous case,  to OpenFace 2.0, and 17
resulting AUs were used as input to a fully connected feed forward neural network,
with two hidden layers of 50 and 25 nodes, using the logistic activation function for all layers.
The F1 score of the trained network is 0.831.

Also in this case the truth space  ${\cal C}_5$ was used.
This means that, with 17 AUs as inputs, 
the size of the search space for a solver is
$6^{17}$, i.e., more than $10^{13}$.
The weighted conditional knowledge base
associated to the network contains 
2\,201
weighted typicality inclusions.
The version of the solver in \cite{arXiv2023}
based on weight constraints and order encoding
was used.
The scalability results in \cite{arXiv2023} 
for synthetic knowledge bases
are consistent with the
theoretical complexity results, showing
that there are solved problem instances as well as
unsolved ones (within a 30 minutes timeout) for search spaces with sizes from $10^7$ to $10^{80}$, and KBs containing
500 to 40\,000 weighted inclusions.

Consider the formulae:
\begin{equation}
\label{F1}
\tip(happiness) \sqsubseteq  \mathit{au1} \sqcup \mathit{au6} \sqcup \mathit{au12} \sqcup \mathit{au14}  \geq k/5
\end{equation}
\begin{equation}
\label{F2}
\tip(happiness) \sqsubseteq  \mathit{au6} \sqcup \mathit{au12} \geq k/5
\end{equation}

Model checking, applied to the test set
for this case (2\,651 individuals with 390 instances of $\tip(happiness)$),
finds that both formulae hold for $k=3$ and
do not hold for $k=4$. As regards entailment:
\begin{itemize}
\item
For (\ref{F1}), the solver 
finds in seconds that it does not hold for $k=4$, and
in minutes that it holds for $k=1$, 
while for $k=2,3$, it does not provide a result in hours.
\item 
On a variant of the experiment, with the same network structure, but using as inputs AU intensities that are not rescaled (so that the AU values are generally lower wrt
the previous case),
the solver finds in seconds that (\ref{F1}) 
does not hold for $k=2$, and in minutes
that it holds for $k=1$.
I.e., in this case the exact separation can be found; note that, in the previous case, the largest $k$ for which the property holds is presumably $k=2$ or $k=3$, where the search space for a counterexample is much less constrained wrt the case $k=1$; such a search
space is relevant both for showing that the property does not hold, if a counterexample is found, and that it does hold, if the non-existence of a counterexample can be inferred. 
\item
for (\ref{F2}),
the property
is found to hold for $k=1$ and not to hold for $k=3$, i.e., for such a value,
a counterexample is found by entailment,
whose search space includes all possible combinations of input vectors,
while it is not found by model checking on the 
(limited) 
test set.
\end{itemize}

The network structure for this experiment 
was chosen to lie in the range considered in the experiments in
\cite{arXiv2023}, even
though a much smaller network with a single hidden
layer of 8 nodes (half of the input nodes) is enough to achieve a similar accuracy
for the classification problem; 
for the resulting knowledge base (with about 150 inclusions) 
the formulae above can be checked in a few seconds even
with a search space of size $6^{17}$.

\subsection{Further considerations}

The entailment approach is definitely more challenging, from the 
computational point of view, than the model-checking one; for the latter, the verification problem is polynomial in time in the size of the domain $\Delta$ and in the size of the formula to be verified. 

The two approaches can be combined, as suggested before, with  model checking providing a guess for the largest value of $k$
such that a formula $\tip(E) \sqsubseteq F \geq k/n$ is entailed.

The entailment approach has been developed for general weighted conditional knowledge bases, which are not required to be acyclic, while  in the experimentation we have considered feedforward networks.
A multilayer network can be seen as a set of weighted defeasible inclusions in a simple description logic (only including boolean concepts).
However, a weighted conditional knowledge base can be more general. It can be defined for several DLs including roles (as it has been done, for instance,  for $\el$ \cite{JELIA2021}, and for $\alc$ in this paper), and it allows for general inclusions axioms and assertions. 
The combination of defeasible inclusions with strict (or fuzzy) inclusions and assertions in a weighted KB allows for the combination of the knowledge acquired from the network and symbolic knowledge in the same formalism.
In the entailment based approach this can be exploited, e.g., by adding constraints on the possible inputs 
through ABox and TBox axioms, e.g., to exclude combinations of input values. 
For example, 
$ au9 \sqcup au10 \sqcup au17 \sqsubseteq \bot \geq 1$,
i.e., imposing that the three AUs have value 0, can be added, assuming
that they are not compatible with $happiness$.
In this case, the properties 
$T(happiness) \sqsubseteq au1 \sqcup au6 \sqcup au12 \sqcup au14 \geq 3 /5$
and
$T(happiness) \sqsubseteq au6 \sqcup au12 \geq 2 /5$
(see section \ref{sec:Entailment})
can indeed be proved in the example (even though hours of computation are needed).

Model checking and entailment are complementary also in the sense
that the limited set of stimuli used for model checking is expected to be
a good sample of the real world, while entailment considers all possible stimuli in the discretized input space, i.e.\ (unless constraints on inputs are used, as described above), it uniformly explores the input space, even though not the whole space of real numbers.
Depending on the purpose of verification, a user may be satisfied with the fact that a formula is verified to hold by model checking, even though counterexamples could be found by entailment.

\section{Conclusions and related work }\label{sec:conclusions}

The paper investigates the relationships between a logic of commonsense reasoning in knowledge representation and multilayer perceptrons.
It develops a fuzzy 
semantics for  weighted knowledge bases with typicality, in which, differently from previous work \cite{JELIA2021,ECSQARU2021}, the typicality operator has a non-crisp interpretation.
For the logic $\alcFt$ we have considered three different closure constructions, thus defining a faithful, a coherent and a $\varphi$-coherent semantics  and  studied the properties of defeasible entailment, proving that the logic satisfies the KLM properties of a preferential consequence relation \cite{whatdoes,Pearl:88} for some choices of fuzzy combination functions. 
We have also considered a finitely many-valued version of the $\varphi$-coherent semantics, the $\varphi_n$-coherent semantics \cite{ICLP22}, and 
proven that it is indeed an approximation of the fuzzy $\varphi$-coherent semantics.  ASP based proof methods for the $\varphi_n$-coherent entailment \cite{arXiv2023} have been exploited in our experimentation.

We have seen that a (fuzzy) multi-preferential interpretation of a trained network can be built from a domain containing a set of input stimuli, and using the activity level of neurons for the stimuli.  
We have proven that such an interpretation is a model of the conditional knowledge base which can be associated to the network, corresponding to a set of weighted defeasible inclusions in a fuzzy description logic.
The logical interpretation of a multilayer network can be used in the verification of properties of the network based on a model checking approach and an entailment-based approach, as experimented on networks recognizing emotions from facial features.

Our semantics builds on fuzzy Description Logics \cite{Straccia05,LukasiewiczStraccia08,BobilloStraccia16}, and we have used fuzzy concepts within a multi-preferential semantics based on semantic closure constructions 
which have been developed in the line of
Lehmann's semantics for lexicographic closure \cite{Lehmann95} and of Kern-Isberner's c-representations \cite{Kern-Isberner01,Kern-Isberner2014}. 
A fuzzy extension of preferential logics has been first studied by Casini and Straccia \cite{CasiniStraccia13_fuzzyRC} for G\"odel logic, based on the Rational closure construction.

The idea of having different preference relations, associated to different typicality operators, has been first explored by Gil \cite{fernandez-gil} to define a multipreference formulation of the description logic $\alctmin$,  
a typicality DL with a minimal model preferential semantics. 
A multi-preferential extension of the rational closure  for $\alc$ and some refinements has been developed by Gliozzi et al. \cite{GliozziAIIA2016,AIJ21}.
The concept-wise multipreference semantics (introduced first in the two-valued case for ranked DL knowledge bases  \cite{TPLP2020}) follows a different route concerning both the definition of preferences, which are associated with concepts, and  the way of combining them. 
In particular, as we have seen in Section \ref{sec:fuzzyalc+T}, in $\alcFt$ the fuzzy interpretation of concepts induces a preference relation over domain elements for each concept, based on the fuzzy combination functions.
An extension of DLs with multiple preferences has also been developed by Britz and Varzinczak \cite{Britz2018,Britz2019} to define defeasible role quantifiers and defeasible role inclusions, by associating multiple preference relations with roles.  A related semantics with multiple preferences has also been proposed in a first-order logic setting by Delgrande and Rantsaudis \cite{Delgrande2020}.

When the preferences associated to concepts are induced by the fuzzy interpretation of concepts, the fuzzy combination functions also provide a notion of {\em preference combination}. 
A related problem of commonsense concept combination has been addressed in a probabilistic extension of the typicality description logic $\alctr$ by Lieto and Pozzato  \cite{Lieto2018}.
In the two valued case, alternative notions of preference combinations have been considered to define a global  
preference relation $<$ from the preferences with respect to single aspects. 
For instance, the multi-preferential semantics for ranked ${\cal EL}_\bot^+$ knowledge bases \cite{TPLP2020} exploits one of the strategies studied in Brewka's framework of basic preference descriptions  \cite{Brewka04},
while an algebraic framework for preference combination in Multi-Relational Contextual Hierarchies has  been developed by Bozzato et al.\ \cite{BozzatoIclp2021}.

In the two valued case, in description logics {\em threshold concepts} have been introduced by Baader et al.\ \cite{BaaderBGFrocos2015}. Graded membership functions $m$  in the semantics assign to a domain element $d$ and a concept $C$ a membership degree $m(d,C)$ in $[0,1]$. The logic is two-valued and the interpretation of concepts and roles is crisp. For instance, a threshold concept  $C_{> 0.8}$   is interpreted as the set of domain elements having a membership  degree in $C$ greater than $0.8$.
Weighted Threshold Operators have as well been introduced in description logics by Porello et al.\ \cite{GallianiEKAW2020}. They are  n-ary operators ${W}^t(C_1: w_1, \ldots, C_n: w_n )$, where the $C_i$ are concepts and the $w_i \in {\mathbb{R}}$ are weights, which compute a weighted sum of their arguments and verify whether it reaches a certain threshold $t$. They are also called perceptron connectives. 
In \cite{Porello2019} an operator ${W}^{max}(C_1: w_1, \ldots, C_n: w_n )$ is also introduced, which selects the set of entities that maximally satisfy a combination of concepts $C_1, \ldots, C_m$. It is proven that the operator can be defined in terms of the  universal modality (in a monotonic DL). 
While the logic $\alcFt$ is monotonic, the notions of  faithful, coherent and $\varphi$-coherent entailment are nonmonotonic, and cannot be encoded in a monotonic description logic (and this is also true for the two-valued case, under the faithful semantics \cite{iclp2021}).

Our semantics, which stems from the combination of (many-valued and fuzzy) DLs semantics \cite{Stoilos05,LukasiewiczStraccia09,BorgwardtPenaloza12},  
and the semantics of preferential logics of 
commonsense reasoning 
\cite{Delgrande:87,Makinson88,Pearl:88,KrausLehmannMagidor:90,Pearl90,whatdoes,BenferhatIJCAI93,BoothParis98, Kern-Isberner01},
has also some relations with Freund's ordered models  for {\em concept representation} \cite{Freund2020}.
Under some respects, our approach can be regarded as a simplification of the ordered model approach (in a many-valued case), as we regard features as concepts and we consider a single (rather than two) preference relation $<_C$ for a concept $C$, which is used for evaluating the degree of typicality of domain elements with respect to $C$, and which is induced by the degree of membership of domain elements in $C$. 
Under these assumptions, simple
multi-preferential structures can be defined and, as we have seen, can be used for providing a semantic interpretation to multilayer networks.
A two-valued version of the concept-wise multi-preferential semantics has  also been considered, e.g., for ranked ${\cal EL}_\bot^+$ knowledge bases \cite{TPLP2020}, for weighted DL knowledge bases \cite{JELIA2021,iclp2021}, and for SOMs \cite{JLC2022}. 
Freund's assumption that the features can be weighted on a finite scale is mitigated in our semantics, by assuming that preferences are well-founded (as usual in the KLM approach \cite{whatdoes}).
However, as we have seen,  restricting to finite values provides an approximation of the fuzzy case.

The correspondence between neural network models and fuzzy systems has been first investigated by Kosko in his seminal work  \cite{Kosko92}.
In his view, ``at each instant the n-vector of neuronal outputs defines a fuzzy unit or a fit vector. Each fit value indicates the degree to which the neuron or element belongs to the n-dimensional fuzzy set.'' In our approach, in a fuzzy interpretation of a multilayer network, each concept (representing a learned category, or simply a unit) is regarded as a fuzzy set over a domain (i.e., a set of input stimuli) 
which is the usual way of viewing concepts in fuzzy description logics \cite{Straccia05,LukasiewiczStraccia08,BobilloStraccia16}, and we have used fuzzy concepts within a multi-preferential semantics based on some semantic closure constructions.
The problem of learning fuzzy rules has been as well investigated in the context of fuzzy description logics \cite{LisiStraccia2015,StracciaMucci2015} 
based on different machine learning approaches.

Much work has been devoted to the combination 
of neural networks and symbolic reasoning 
(e.g., the work by d'Avila Garcez et al.\ \cite{GarcezBG01,GarcezLG2009,GarcezGori2019} and Setzu et al.\ \cite{Guidotti21}), as well as to the definition of new computational models \cite{GarcezGori2020,SerafiniG16,Lukasiewicz2020,PhuocEL21},
and to extensions of logic programming languages
with neural predicates \cite{DeepProbLog18,NeurASP2020}.
Among the earliest 
systems combining logical reasoning and neural learning are the Knowledge-Based Artificial Neural Network (KBANN) \cite{KBANN94}, the Connectionist Inductive Learning and Logic Programming (CILP) \cite{CLIP99}
systems, and Penalty Logic \cite{Pinkas95}, a non-monotonic reasoning formalism used to establish a correspondence with symmetric connectionist networks.
The relationships between normal logic programs and connectionist network have been investigated by Garcez and Gabbay \cite{CLIP99,GarcezBG01}
and by Hitzler et al.\ \cite{HitzlerJAL04}.
None of these approaches provides a semantics of neural networks in terms of concept-wise multi-preferential interpretations with typicality.

The work presented in this paper  opens to the possibility of adopting conditional logics as a basis for neuro-symbolic integration, e.g.,  
by learning the weights of a conditional knowledge base from empirical data, 
and combining the defeasible inclusions extracted from a neural network with other defeasible or strict inclusions for inference.

Using a multi-preferential logic for the verification of typicality properties of a neural network by model-checking is a general ({\em model agnostic}) approach. 
It can be used for  SOMs, as in \cite{JLC2022}, by exploiting a notion of {\em distance} of a stimulus from a category to define a preferential structure, as well as for MLPs, by exploiting units activity to build a fuzzy preferential interpretation. Given the simplicity of the approach,
a similar construction
can be adapted to other neural network models and learning approaches. 

Both the model-checking approach and the entailment-based approach are {\em global} approaches to explanation for neural networks (see, e.g., \cite{Guidotti21} for the notions of local and global approaches), as they consider the behavior of the network over a set $\Delta$ of  input stimuli. 
Indeed, the evaluation of typicality inclusions considers all the individuals in the domain to establish preference relations among them, with respect to different aspects. 
However, properties of single individuals can as well be verified (by instance checking, in DL terminology). 
Whether this approach can as well be considered for counterfactual reasoning has still to be investigated.

The model-checking approach does not require to consider the activity of all units, but only of the units involved in the property to be verified. In the entailment-based approach, on the other hand, all units and network parameters are considered,
which limits the scalability of the approach,
consistently with the complexity results.
Whether it is possible to extend the logical encoding of MLPs as weighted KBs  to other neural network models is a subject for future investigation. 
The development of a temporal extension  
of this formalism to capture the transient behavior of MLPs is also an interesting direction to extend this work.

\section*{Acknowledgments}
This work was partially supported by MUR and GNCS-INdAM. 
Mario Alviano was partially supported by Italian Ministry of Research (MUR) 
under PNRR project FAIR ``Future AI Research'', CUP H23C22000860006 and
by the LAIA lab (part of the SILA labs).

\newpage

\begin{appendix}

\section{Proofs of Propositions \ref{prop:KLM_properties} and \ref{prop:RM}}

{\bf Proposition \ref{prop:KLM_properties}}
{\em Under the choice of combination functions as in G\"odel logic, 
any $\alcFt$ interpretation $I=\langle \Delta, \cdot^I \rangle $
satisfies the postulates $\mathit{(REFL')}$, $\mathit{(LLE')}$, $\mathit{(RW')}$, $\mathit{(AND')}$, $\mathit{(OR')}$ and $\mathit{(CM')}$.
}

\begin{proof}
Let $I= \langle \Delta, \cdot^I \rangle$ be an $\alcFt$ interpretation, 
where the t-norm, s-norm, implication function and negation functions are as in G\"odel logic. To prove that $I$ satisfies the properties 
$(REFL'), (LLE'), (RW'),(AND')$, $(OR')$ and $(CM')$,  
when the t-norm, s-norm and implication function are as in G\"odel logic,
while for negation we adopt standard involutive negation.
We proceed by cases.

\medskip

\noindent
$ \bf -  (REFL')$ \ to prove that $\tip(C) \sqsubseteq C \geq 1$ is satisfied in $I$, we have to prove that
$\inf_{x \in \Delta} (\tip(C))^I(x) \rhd C^I(x) \geq 1$.

Let us prove that $\mbox{for all } x \in \Delta$, 
$(\tip(C))^I(x) \rhd C^I(x) \geq 1 $.

We consider two cases: $(\tip(C))^I (x) =0$ (i.e., $x$ is not a typical $C$-element) and $(\tip(C))^I (x) >0$ (i.e., $x$ is a typical $C$-element).

If $(\tip(C))^I (x)=0$, $(\tip(C))^I(x) \rhd C^I(x) = 0  \rhd C^I(x) =1 $, and the thesis holds trivially.

If $(\tip(C))^I (x) >0$, by definition $(\tip(C))^I (x) = C^I(x)$. Again, $(\tip(C))^I(x) \rhd C^I(x) =1 $, and the thesis holds.

\medskip

\noindent
$ \bf -  (LLE')$ \ 
Assume that $\models A \equiv B$, i.e., axioms $A \sqsubseteq B \geq 1$, $B \sqsubseteq A \geq 1$ are valid in fuzzy $\alc$ and that $\tip(A) \sqsubseteq C \geq k $ is satisfied in $I$.
We prove that $\tip(B) \sqsubseteq C \geq 1$ is satisfied in $I$, that is $(\tip(B) \sqsubseteq C)^I \geq 1$.

From the validity of $A \sqsubseteq B \geq 1$ and $B \sqsubseteq A \geq 1$,  
$\inf_{x \in \Delta} A^I(x) \rhd B^I(x) \geq 1$ and $\inf_{x \in \Delta} B^I(x) \rhd A^I(x) \geq 1$. 
Hence,
\begin{equation}\label{eq:LLC}
\mbox{for all } x \in \Delta, \; A^I(x) \rhd B^I(x) \geq 1 \mbox{ and } B^I(x) \rhd A^I(x) \geq 1
\end{equation}
 
\noindent
This implies that: for all $x \in \Delta$, $A^I(x) \leq B^I(x)$ and $B^I(x) \leq A^I(x)$, i.e., $A^I(x) = B^I(x)$ for all $x \in \Delta$.
Therefore, the preference relations $<_A$ and $<_B$ must be the same and also $A^I_{>0}= B^I_{>0}$.
Hence, $\tip(A)^I(x) = \tip(B)^I(x)$ for all $x \in \Delta$, and from $(\tip(A) \sqsubseteq C)^I \geq 1 $, it follows that $(\tip(B) \sqsubseteq C)^I \geq 1 $, that is,  $\tip(B) \sqsubseteq C \geq 1$ is satisfied in $I$.

\medskip

\noindent
$ \bf  - (RW')$ \  
Assume that axiom $C \sqsubseteq D \geq 1$ is valid in fuzzy $\alc$.
Hence, it holds that
$\inf_{x \in \Delta} C^I(x) \rhd D^I(x) \geq 1$ and $\mbox{for all } x \in \Delta, \; C^I(x) \rhd D^I(x) \geq 1 $.

As we have seen above, this implies that: for all $x \in \Delta$, $C^I(x) \leq D^I(x)$. 

Let us assume that $\tip(A) \sqsubseteq C \geq 1$ is satisfied in $I$, i.e., $\inf_{x \in \Delta} (\tip(A))^I(x) \rhd C^I(x) \geq 1$ and,  $\mbox{for all } x \in \Delta$, 
$(\tip(A))^I(x) \rhd C^I(x) \geq 1 $.

By monotonicity of $\rhd$, $1 \leq (\tip(A))^I(x) \rhd C^I(x) \leq   (\tip(A))^I(x) \rhd D^I(x) $.
Hence, $\mbox{for all } x \in \Delta$, 
$(\tip(A))^I(x) \rhd D^I(x) \geq 1 $, so that $\tip(A) \sqsubseteq D \geq 1$ is satisfied in $I$.

\medskip

\noindent
$ \bf - (AND')$ \  
Let us assume that $\tip(A) \sqsubseteq C \geq 1$ and $\tip(A) \sqsubseteq D \geq 1$ are satisfied in $I$, i.e.,
$\inf_{x \in \Delta} (\tip(A))^I(x) \rhd C^I(x) \geq 1$ and $\inf_{x \in \Delta} (\tip(A))^I(x) \rhd D^I(x) \geq 1$.
Then,
For all $x \in \Delta$, 
$(\tip(A))^I(x) \rhd C^I(x) \geq 1 $ and $(\tip(A))^I(x) \rhd D^I(x) \geq 1 $.

We prove that $x \in \Delta$, $(\tip(A))^I(x) \rhd (C \sqcap D)^I(x) \geq 1 $, from which $(\tip(A) \sqsubseteq C \sqcap D)^I \geq 1$ follows.

If $(\tip(A))^I(x) \rhd C ^I(x) \geq 1 $ holds, then  (a)  $(\tip(A))^I(x) \leq C^I(x)$ or (b) $C^I(x) \geq 1 $.
Note that, if (b) holds, (a) must hold as well. Hence, if $(\tip(A))^I(x) \rhd C ^I(x) \geq 1 $ holds,  $(\tip(A))^I(x) \leq C^I(x)$ also holds.

Similarly, from $(\tip(A))^I(x) \rhd D ^I(x) \geq 1 $, it follows that $(\tip(A))^I(x) \leq D^I(x)$ holds. 
 
Therefore, for any $x \in \Delta$, both  $(\tip(A))^I(x) \leq C^I(x)$ and   $(\tip(A))^I(x) \leq D^I(x)$ hold. It follows that $(\tip(A))^I(x) \leq min\{C^I(x), D^I(x)\} =(C \sqcap D)^I(x)$ holds
and, hence, $(\tip(A))^I(x) \rhd (C \sqcap D)^I(x) \geq 1 $ holds.

\medskip
\noindent
$ \bf - (OR')$ 
Assume $\tip(A) \sqsubseteq C \geq 1$ and that  $\tip(B) \sqsubseteq C \geq 1$ are satisfied in $I$.
Then, $\inf_{x \in \Delta} (\tip(A))^I(x) \rhd C^I(x) \geq 1$ and $\inf_{x \in \Delta} (\tip(B))^I(x) \rhd C^I(x) \geq 1$.
Hence,
for all $x \in \Delta$, 
$(\tip(A))^I(x) \rhd C^I(x) \geq 1 $ and $(\tip(B))^I(x) \rhd C^I(x) \geq 1 $.

To prove that $\tip(A \sqcup B) \sqsubseteq C \geq 1$, we prove that, for all $x \in \Delta$, 
$(\tip(A \sqcup B))^I(x) \rhd C^I(x) \geq 1 $. 

If $(\tip(A \sqcup B)^I(x)=0$, the thesis follows trivially.

If $(\tip(A \sqcup B))^I(x)>0$, $x$ is a  typical $A \sqcup B$-element. 
Then there is no $y \in \Delta$ such that $(A \sqcup B)^I(y) > (A \sqcup B)^I(x)$.

It can be proven that, when $x$ is a typical $A \sqcup B$-element, $x$ is also a typical $A$-element or a typical $B$-element. 

Given that $(\tip(A \sqcup B))^I(x) =(A \sqcup B)^I(x)= max\{ A^I(x), B^I(x)\}$, let us 
assume $max\{ A^I(x), B^I(x)\} = B^I(x)$.
Then, for all $y \in \Delta$,  $max\{ A^I(y), B^I(y)\} \leq B^I(x)$,
and $B^I(y) \leq B^I(x)$.
Hence, there is no $y \in \Delta$ such that $ B^I(y) > B^I(x)$, and $x$ is a typical $B$ element. 
Furthermore, $(\tip(B))^I(x)= B^I(x)= (\tip(A \sqcup B))^I(x)$).

From the hypothesis, we know that $(\tip(B))^I(x) \rhd C^I(x) \geq 1 $; hence, $(\tip(B))^I(x) \leq C^I(x) $.
It follows that $(\tip(A \sqcup B))^I(x) = (\tip(B))^I(x)  \leq C^I(x) $,
and then $(\tip(A \sqcup B))^I(x) \rhd C^I(x) \geq 1 $.

The case where $max\{ A^I(x), B^I(x)\} = A(x)$ is similar.

\medskip
\noindent
$ \bf - (CM')$  
Assume $\tip(A) \sqsubseteq D \geq 1$ and that  $\tip(A) \sqsubseteq C  \geq 1$ are satisfied in $I$.
Then, $\inf_{x \in \Delta} (\tip(A))^I(x) \rhd D^I(x)  \geq 1$ and $\inf_{x \in \Delta} (\tip(A))^I(x) \rhd C^I(x) \geq 1$.
Hence,
for all $x \in \Delta$, 
$(\tip(A))^I(x) \rhd D^I(x)  \geq 1 $ and $(\tip(A))^I(x) \rhd C^I(x)  \geq 1 $ .

To prove that $\tip(A \sqcap D) \sqsubseteq C  \geq 1$ is satisfied in $I$, we prove that, for all $x \in \Delta$, 
$(\tip(A \sqcap D))^I(x) \rhd C^I(x)  \geq 1 $.

If $(\tip(A \sqcap D))^I(x) =0$ the thesis holds trivially.

If $(\tip(A \sqcap D))^I(x)>0$, $x$ is a  typical $A \sqcap D$-element. 
Then, $(\tip(A \sqcap D))^I(x) =(A \sqcap D)^I(x)= min\{ A^I(x), D^I(x)\} > 0$. Also, $A^I(x) >0$ and $D^I(x)>0$.

We prove that $x$ is a typical $A$-element. By contradiction, if $x$ is not a typical $A$-element, there is a $y \in \Delta$ such that 
 $y$ is a typical $A$-element and $A^I(y) > A^I(x)$.
As $\tip(A) \sqsubseteq D \geq 1$, $(\tip(A))^I(y) \rhd D^I(y)  \geq 1 $, and then
 $(\tip(A))^I(y) \leq D^I(y)$. But $(\tip(A))^I(y)= A^I(y)$ (as  $y$ is a typical $A$-element) , hence $A^I(y) \leq D^I(y)$. 
 
 Therefore, $min\{  A^I(y), D^I(y) \} = A^I(y) > A^I(x) \geq min\{  A^I(x), D^I(x) \} $. Then, $(A \sqcap D)^I(y) > (A \sqcap D)^I(x) $,
 which contradicts the hypothesis that $x$ is a typical $A \sqcap D$-element.
 Therefore, $x$ must be a typical $A$-element.
 
 As  $\tip(A) \sqsubseteq C  \geq 1$ and $\tip(A) \sqsubseteq D  \geq 1$, 
$(\tip(A))^I(x) \leq C ^I(x) $ and $(\tip(A))^I(x) \leq D ^I(x) $ hold.
Furthermore, as $(\tip(A))^I(x)=A^I(x)$,  $A^I(x) \leq C ^I(x) $ and $A^I(x) \leq D ^I(x) $ hold.
Then, $(A \sqcap D)^I(x) = min\{  A^I(x), D^I(x) \} = A^I(x)$.
Thus, 
  $(\tip (A \sqcap D))^I(x) =(A \sqcap D)^I(x) = A^I(x) \leq C^I(x)$, and the thesis follows.

\end{proof}

\noindent
{\bf Proposition \ref{prop:RM}}.
{\em For the choice of combination functions as in G\"odel logic, $\mathit{(RM')}$ does not hold in $\alcFt$
(and the same with standard involutive negation.}

\begin{proof}
Consider a KB $K$ such that the ABox ${\cal A}$ contains the following assertions:

$A(a)\leq 0.8$, $A(a)\geq 0.8$ $B(a)\leq 0.3$, $B(a)\geq 0.3$, $C(a)\leq 0.9$, $C(a)\geq 0.9$;

$A(b)\leq 0.5$, $A(b)\geq 0.5$,  $B(b)\leq 0.6$, $B(b)\geq 0.6$, $C(b)\leq 0.4$, $C(b)\geq 0.4$

\noindent
and the TBox ${\cal T}$ contain the axiom $\tip(A) \sqsubseteq C \geq 1$.

Clearly $K$ entails $\tip(A) \sqsubseteq C \geq 1$.
We show that $K$ does not entail $\tip(A) \sqsubseteq \neg B \geq 1$.
We define an $\alcFt$ interpretation  $I=\langle \Delta, \cdot^I \rangle $ which is a model of $K$, but falsifies $\tip(A) \sqsubseteq \neg B \geq 1$.

Let $I=\langle \Delta, \cdot^I \rangle $ be  such that $\Delta=\{x,z\}$ and, for concept names $A$, $B$, $C$,
\begin{quote}
$A^I(x)=0.8$, $B^I(x)=0.3$, $C^I(x)=0.9$

$A^I(z)=0.5$, $B^I(z)=0.6$, $C^I(z)=0.4$
\end{quote}
Hence, $x$ is a typical $A$ element, and the only one. $\tip(A)^I(x)=0.8$ and $\tip(A)^I(x) \rhd C^I(x)=1$. Hence,  $\tip(A) \sqsubseteq C \geq 1$ is satisfied in $I$. Clearly, all the assertions in ABox ${\cal A}$ are also satisfied in $I$, by letting $a^I=x$ and $b^I=z$. $I$ is an $\alcFt$ model of $K$.

$\tip(A) \sqsubseteq \neg B \geq 1$ is not satisfied in $I$, as $\tip(A)^I(x)=0.8$ and $(\neg B)^I(x)=0$ (using the negation function in G\"odel logic), 
and $\tip(A)^I(x) \rhd (\neg B)^I(x)=0$. Therefore, $\tip(A) \sqsubseteq \neg B \geq 1$ is not entailed from $K$.\footnote{Similarly, using standard involutive negation, $(\neg B)^I(x)=1-0.3=0.7$,  $\tip(A)^I(x) \rhd (\neg B)^I(x)< 1$, and $\tip(A) \sqsubseteq \neg B \geq 1$ is not satisfied in $I$ as well, and $\tip(A) \sqsubseteq \neg B \geq 1$ is not entailed from $K$.}

By (RM') we would conclude that $\tip(A \sqcap B) \sqsubseteq C \geq 1$ should be entailed from $K$, but this is not true, as the model $I$ of $K$ falsifies $\tip(A \sqcap B) \sqsubseteq C \geq 1$. In fact $z$ is the only typical $\tip(A \sqcap B)$ element in $I$
and $\tip(A \sqcap B)^I(z)=0.5$. However, $\tip(A \sqcap B)^I(z) \rhd C^I(z) =0.4 <1$.  
\end{proof}

\end{appendix}

\end{document}